\documentclass[journal]{IEEEtran}

\usepackage{cite}
\ifCLASSINFOpdf
  \usepackage[pdftex]{graphicx}
  \usepackage{float}
  \graphicspath{{./pdf/}{./jpeg/}}
  \DeclareGraphicsExtensions{.pdf,.jpeg,.png}
\else
\fi

\usepackage{epsfig}
\usepackage{graphicx}
\graphicspath{{./pdf/}{./jpeg/}}
\DeclareGraphicsExtensions{.pdf,.jpeg,.png,.tif}

\usepackage[cmex10]{amsmath}
\usepackage{amssymb}
\usepackage{array}
\usepackage{makecell}
\usepackage{tabularx}
\usepackage{multicol}
\usepackage{multirow, makecell}
\usepackage{booktabs}
\usepackage{threeparttable}
\usepackage{arydshln}

\usepackage{cite}
\usepackage{times}
\usepackage{psfrag}
\usepackage{stfloats}

\newcolumntype{Y}{>{\centering\arraybackslash}X}

\usepackage{fixltx2e}

\ifCLASSOPTIONcaptionsoff
 \usepackage[nomarkers]{endfloat}
 \let\MYoriglatexcaption\caption
 \renewcommand{\caption}[2][\relax]{\MYoriglatexcaption[#2]{#2}}
\fi

\usepackage{url}

\begin{document}

\title{Crowded Scene Analysis: A Survey}

\author{Teng~Li, 
        Huan~Chang, 
        Meng~Wang,
        Bingbing~Ni,
        Richang~Hong,
        and Shuicheng~Yan, ~\IEEEmembership{Senior Member,~IEEE}
\thanks{T. Li and H. Chang are with Anhui University, Hefei 230601, P. R. China (email: tenglwy@gmail.com; changhuan@ahu.edu.cn); M. Wang and R. Hong are with Hefei University of Technology (email: eric.mengwang@gmail.com; hongrc.hfut@gmail.com);
B. Ni is with Advanced Digital Sciences Center, 1 Fusionoplis Way, \#08-10 Connexis North Tower, 138632, Singapore (email: bingbing.ni@adsc.com.sg); S. Yan is with National University of Singapore, Singapore 117576 (email: eleyans@nus.edu.sg).}
\thanks{This work is supported by the National Natural Science Foundation (NSF) of China (No. 61300056, 61272393, 61322201), the Anhui Provincial Natural Science Foundation of China (No. 1408085QF118) and the Open Project Program of the National Laboratory of Pattern Recognition (NLPR) (No. 201306282).
This work is also supported in part by the National 973 Program of China (No. 2014CB347600), and a research grant for the Human Sixth Sense Programme at the Advanced Digital Sciences Center from Singapore’s Agency for Science, Technology and Research (A*STAR).}
\thanks{${}$Copyright (c) 2014 IEEE. Personal use of this material is permitted. However, permission to use this material for any other purposes must be obtained from the IEEE by sending a request to pubs-permissions@ieee.org.}
}

\markboth{IEEE TRANSACTIONS ON XXX, ~Vol.~X, No.~XX, ~Month~Year}{}


\maketitle

\begin{abstract}
    Automated scene analysis has been a topic of great interest in computer vision and cognitive science. Recently, with the growth of crowd phenomena in the real world, crowded scene analysis has attracted much attention. However, the visual occlusions and ambiguities in crowded scenes, as well as the complex behaviors and scene semantics, make the analysis a challenging task. In the past few years, an increasing number of works on crowded scene analysis have been reported, covering different aspects including crowd motion pattern learning, crowd behavior and activity analysis, and anomaly detection in crowds. This paper surveys the state-of-the-art techniques on this topic. We first provide the background knowledge and the available features related to crowded scenes. Then, existing models, popular algorithms, evaluation protocols, as well as system performance are provided corresponding to different aspects of crowded scene analysis. We also outline the available datasets for performance evaluation. Finally, some research problems and promising future directions are presented with discussions.

\end{abstract}


\ifCLASSOPTIONpeerreview
\begin{center} \bfseries EDICS Category: 3-BBND \end{center}
\fi
%
\IEEEpeerreviewmaketitle

\section{Introduction} \label{sec:introduction}

\begin{figure}[htb]
        \centering
        \includegraphics[width = \linewidth]{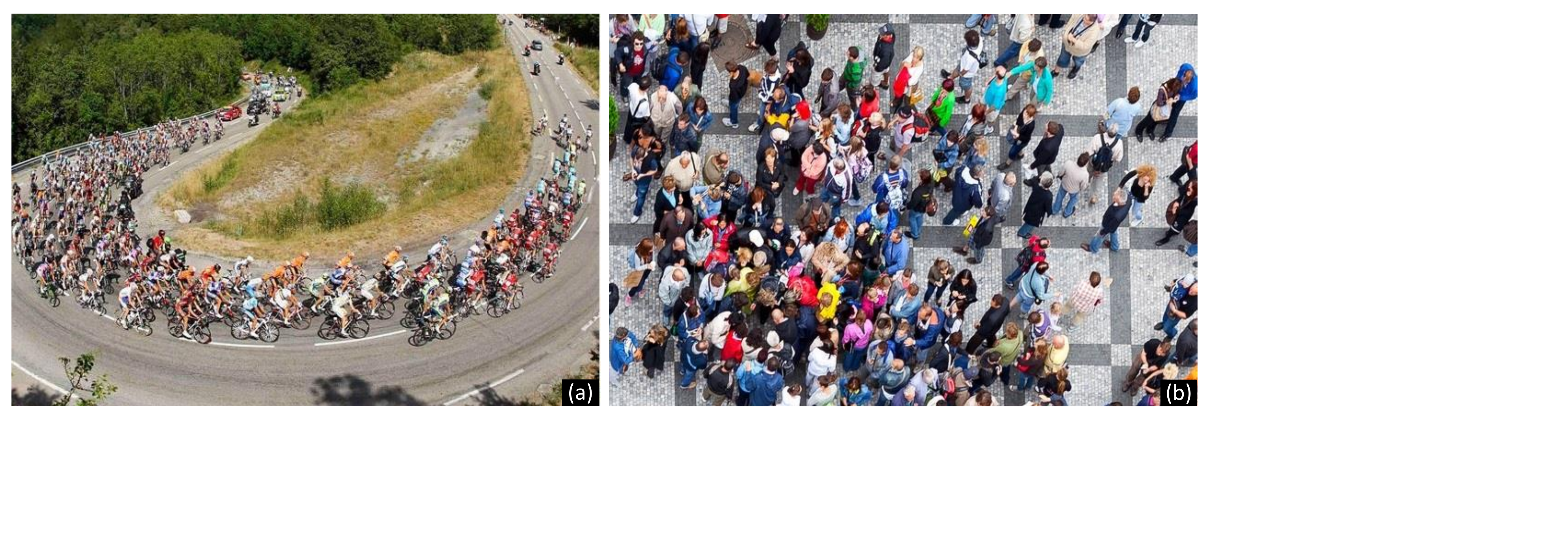}
        \caption{Examples of crowded scenes: (a) a stage of the tour de france (structured scene); (b) people gathered in a square (unstructured scene).}
        \label{fig:fig1}
    \end{figure}

    \IEEEPARstart{W}{ith} the increase of population and diversity of human activities, crowded scenes have been more frequent in the real world than ever. It brings enormous challenges to public management, security or safety. Some examples of crowded scenes are shown in Figure \ref{fig:fig1}.
    Humans have the ability to extract useful information of behavior patterns in the surveillance area, monitor the scene for abnormal situations in real time, and provide the potential for immediate response \cite{hospedales2012video}.
    However, psychophysical research indicates that there are severe limitations in their ability to monitor simultaneous signals \cite{sulman2008how}. Extremely crowded scenes require monitoring an excessive number of individuals and their activities, which is a significant challenge even for a human observer.

    In the past decade, automated scene understanding or analysis has already attracted much research attention in the computer vision community \cite{saleemi2010scene,44,48,49,50,54}.
    One important application is intelligent surveillance in replace of the traditional passive video surveillance. Although many algorithms have been developed to track, recognize and understand the behaviors of various objects in video~\cite{36}, they were mainly designed for common scenes with a low density of population \cite{48,49,64}.
When it comes to crowded scenes, the problems can not be handled well, since the large number of individuals involved not only cause the detection and tracking fail, but also greatly increase computational complexity.
    Under such circumstance, crowded scene analysis as a unique topic, is specifically addressed. Driven by the practical demand, it is becoming an important research direction and has already attracted lots of efforts \cite{1,16,17,27,35,36,37}. The opportunity for such study has never been better.

As noted in \cite{saleemi2010scene,44}, scene understanding may refer to scene layout (locating roads, buildings, sidewalks) \cite{14}, motion patterns (vehicles turning, pedestrian crossing) \cite{saleemi2010scene,44,45,46} and scene status (crowd congestion, split, merge, etc.) \cite{27,47}. In this paper, combined with previous studies, we will elaborate the key aspects of crowded scene analysis in automated video surveillance.

\subsection{Real-World Applications}
    Research of crowded scene analysis could lead to a lot of critical applications.

\subsubsection{Visual Surveillance}
    Many places of security interests such as railway station and shopping mall are very crowded. Conventional surveillance system may fail for high density of objects, regarding both accuracy and computation. We can leverage the results of crowd behavior analysis to crowd flux statistics and congestion analysis \cite{47,51}, anomaly detection and alarming \cite{1,16,17,25}, etc.

\subsubsection{Crowd Management}
    In mass gatherings such as music festivals and sports events, crowded scene analysis can be used to develop crowd management strategies and assist the movement of the crowd or individuals, to avoid the crowd disasters and ensure the public safety \cite{70,71}.

\subsubsection{Public Space Design}
    The analysis of crowd dynamics and its relevant findings \cite{142,143} can provide some guidelines for public space design, and therefore increase the efficiency and safety of train stations, airport terminals, theaters, public buildings, and mass events in the future.

\subsubsection{Entertainment}
    With the in-depth understanding of crowd phenomena, the establishment of mathematical models can provide more accurate simulation, which can be used in computer games, film and television industries \cite{145}.
    Some recent works have been proposed to synthesize crowd videos with realistic microscale behavior~\cite{flagg2013video}.

\subsection{Problems and Motivations}
    Video analysis and scene understanding usually involve object detection, tracking and behavior recognition \cite{48,49}.
    For crowded scenes, due to extreme clutters, severe occlusions and ambiguities, the conventional methods without special considerations are not appropriate.
    As Ali pointed out \cite{4}, the mechanics of human crowds are complex as a crowd exhibits both dynamics and psychological characteristics, which are often goal directed. This makes it very challenging to figure out an appropriate level of granularity to model the dynamics of a crowd. Another challenge in crowded scene analysis is that the specific crowd behaviors needed to be detected and classified may be both rare and subtle \cite{33}, and in most surveillance scenarios, these behaviors have few examples to learn.

    These challenges have partially drawn attention at some recent conferences, and several relevant scientific papers have also been published in academic journals. In this paper, we try to explore those problems with a comprehensive review and general discussions. The state-of-the-art technical advances in crowded scene analysis will be covered.
    Feature extraction, segmentation and model learning are considered as core problems addressed in visual behavior analysis \cite{34}. We will discuss the methods in crowded scene analysis regarding these basic issues.

    It is noted that survey papers \cite{35,36,38,thida2013aliteraturereview} relevant to the topic of crowd analysis have been written in the past few years. Zhan \emph{et al.} \cite{35} presented a survey in 2008 on crowd analysis methods in computer vision. They covered the techniques of crowd density estimation, pedestrian/crowd recognition and crowd tracking. They paid much attention on the perspectives from other research disciplines, such as sociology, psychology and computer graphics, to the computer vision approach. The paper \cite{36} by Junior \emph{et al.} in 2010 also presented a survey on a wide range of computer vision techniques for crowd analysis, covering people tracking, crowd density estimation, event detection, validation and simulation. They devoted large sections to reporting how related the areas of computer vision and computer graphics should be to deal with challenges in crowd analysis. Later in 2011, Sjarif \emph{et al.} \cite{38} wrote a survey with more emphasis on abnormal behaviors detection in crowded scenes.
    Recently, Thida \emph{et al.} \cite{thida2013aliteraturereview} gave a general review on crowd video analysis, providing some valuable summarizations. But still many recent important works have been missed, and the descriptions of methods were rather brief.

    In this study, we seek to create a more focused review of recent publications on high-level crowded scene understanding, related to motion and behavior analysis in crowd videos. Several important recent works from 2010 to now will be covered and compared, which have not yet been surveyed previously. In order to better elaborate this topic, we divide it into three subtopics according to the purposes of task: motion pattern segmentation, crowd behavior recognition and anomaly detection. It is indicated that these three aspects are closely related to each other in crowded scene analysis \cite{134}.
    Differently from previous survey such as \cite{35}, crowd counting or density estimation, a closely related topic~\cite{chen2012feature,ma2013crossing,conte2013counting}, is not covered in this survey.
    We intend to cover the area related to the analysis of behaviors and activities in crowd videos, which is broad enough. The reviewed methods are all about the behaviors or activities, and are usually based on the motion features, while in crowd counting, static visual features can be important.

\begin{figure}[tbp]
        \centering
        \includegraphics[width = \linewidth]{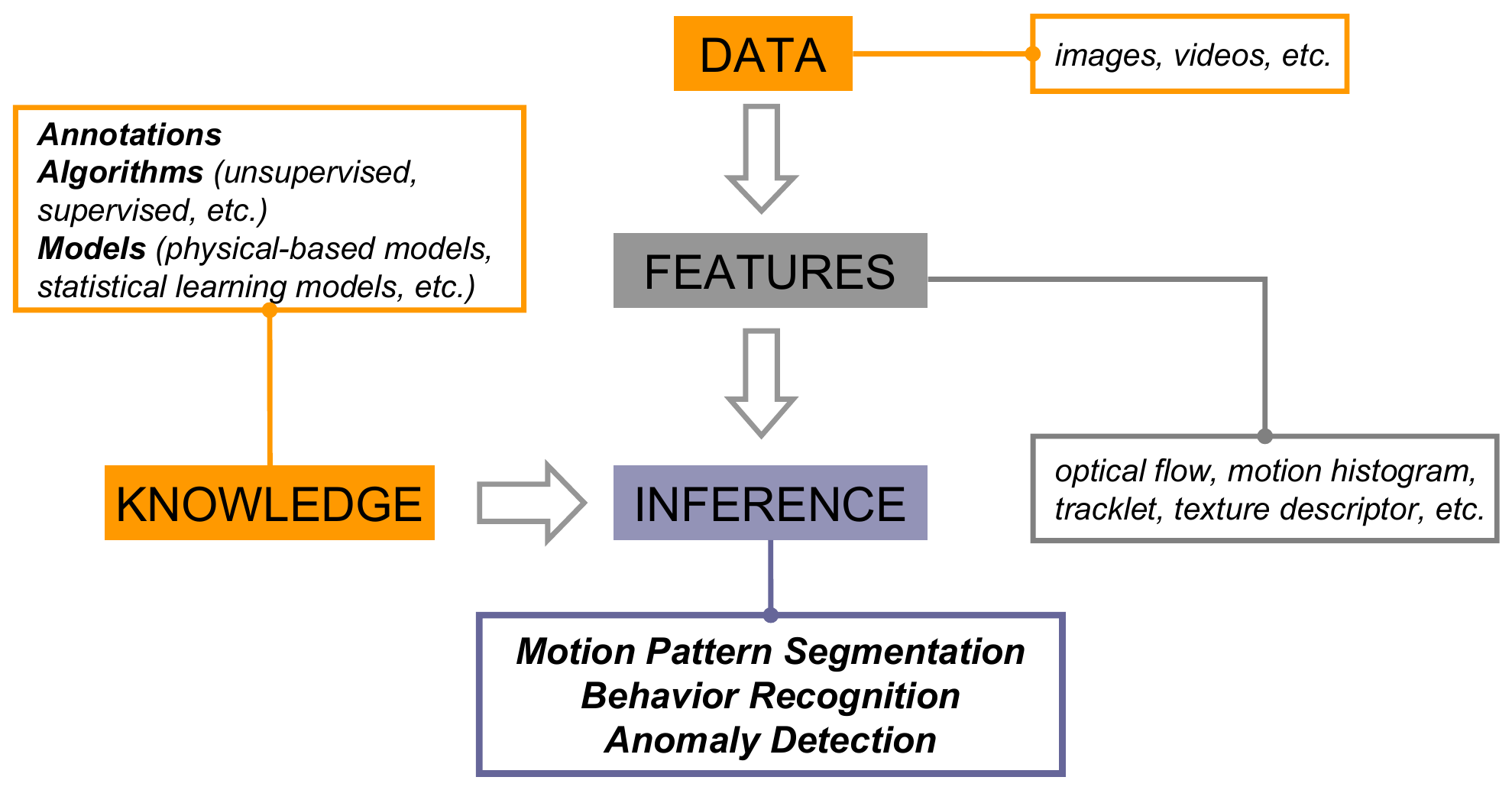}
        \caption{General structure of crowded scene analysis, also illustrating the main issues discussed in this survey.}
        \label{fig:fig3}
    \end{figure}

    As an indispensable basis for each crowded scene analysis method, feature representation is discussed and summarized  separately before elaboration of the three subtopics. Feature representation can be shared by different methods. To clearly and properly situate the problem at hand for the readers, we still follow the task-orientated way to categorize the methods into three subtopics. Besides, some background knowledge and available physical crowd models are provided beforehand. They could be utilized in the analysis methods.
    Figure \ref{fig:fig3} illustrates the diagram of crowded scene analysis. It also reveals what this paper is going to elaborate and explain.

\subsection{Organization of The Paper}
    The remainder of this paper is organized as follows. Section \ref{sec:theory} describes some background knowledge and physical models that could be utilized in crowded scene analysis. In Section \ref{sec:feature}, we introduce the features commonly used in the literature. In Section \ref{sec:segmentation}, the concept of motion pattern segmentation and the relevant approaches are detailed. Section \ref{sec:recognition} provides a detailed review on the topic of crowd behavior recognition, and the approaches for anomaly detection are elaborated in Section \ref{sec:anomaly}. Section \ref{sec:dataset} provides the detailed information of frequently used datasets for crowded scene analysis. We conclude the paper in Section \ref{sec:conclusion} and end this review by providing some promising future directions.

\section{Knowledge of The Crowd} \label{sec:theory}
    Crowded scenes can be divided into two categories according to the motion of the crowd\cite{19}: \emph{structured} and \emph{unstructured}. In the structured crowded scenes, the crowd moves coherently in a common direction, the motion direction does not vary frequently, and each spatial location of the scene contains only one main crowd behavior over the time. The unstructured crowded scenes represent the scenes with chaotic or random crowd motion, where participants move in different directions at different times, and each spatial location contains multiple crowd behaviors~\cite{37}.
    Figure \ref{fig:fig1}(a) is structured crowded scenes, while Figure \ref{fig:fig1}(b) is unstructured scenes. Obviously, they have different dynamic and visual characteristics.

    The crowd has been defined as ``a large group of individuals in the same physical environment, sharing a common goal" \cite{musse1997model}. It can be viewed hierarchically: individuals are collected into groups, and the resulting groups are collected into a crowd, with a set of motivations and basic rules \cite{andrade2005simulation}.
    This representation permits a flexible analysis of a large spectrum of crowd densities and complicated behaviors.

    The analysis of crowd can be conducted at macroscopic or microscopic levels. At the macroscopic level, we are interested in the global motions of a mass of people, without concerning the movements of any individual; at the microscopic level, we concern the movements of each individual pedestrian and do analyze based on the collective information of them.

    The analysis of crowded scenes could involve the knowledge from both vision area and crowd dynamics. Except for the vision algorithms usually adopted in conventional scene analysis, the physical models from crowd dynamics could also be utilized.
    In the below we will introduce some available knowledge such as the models from crowd dynamics, as well as their application in crowded scene analysis.

\subsection{Models in Crowd Dynamics}
    Crowd dynamics have been studied intensively for more than 40 years. It can be considered as the study of how and where crowds form and move above the critical density level \cite{still2000crowd}, and how individuals in the crowd interact with each other to influence the crowd status. In the past, studies were mainly conducted in order to support planning of urban infrastructure, e.g., building entrances and corridors \cite{brambilla1977pedestrians}.

    There are two major approaches for computational modeling of crowd behavior: continuum-based approach and agent-based approach.
    In crowd simulation, both of them have been frequently used in to reproduce crowd phenomena.
    Continuum-based approach works better at the macroscopic level for medium and high density crowds, while the agent-based approach is more suitable for low density crowds at the microscopic level, where the movement of each individual pedestrian is concerned.

    For the continuum-based approach, the crowd is treated as a physical fluid with particles, thus a lot of analytical methods from statistical mechanics and thermodynamics are introduced \cite{hughes2002continuum, treuille2006continuum}.
    Hughes \emph{et al.} \cite{hughes2002continuum} have developed a model representing pedestrians as a continuous density field, and have presented a pair of elegant partial differential equations describing the crowd dynamics.
    Moreover, a real-time crowd model based on continuum dynamics has been presented in \cite{treuille2006continuum}. It could yield a set of dynamic potentials and velocity fields to guide the individuals' motions.

\begin{figure}[tbp]
        \centering
        \includegraphics[width = 0.9\linewidth]{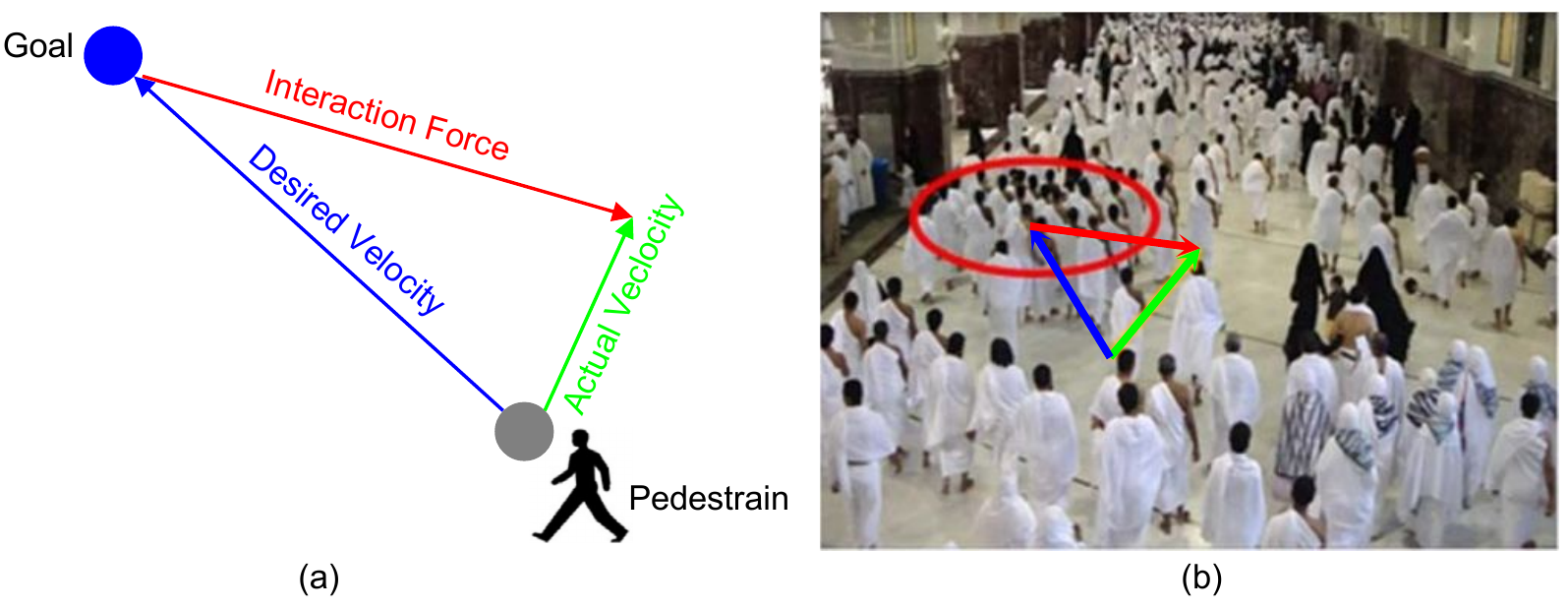}
        \caption{An exemplar visualization of the forces and velocities in the SFM.}
        \label{fig:fig11}
    \end{figure}

    For the agent-based approach, individuals in the crowd are considered as autonomous agents which actively sense the environment and make decisions according to some predefined rules \cite{helbing1995social, still2000crowd}.
    Following this style, the social force model (SFM), first proposed by Helbing \emph{et al.} \cite{helbing1995social}, has been proven to be capable of reproducing specific crowd phenomena. The assumption is that the interaction force between pedestrians is a significant feature for analyzing crowd behaviors. The SFM can be formulated as:
    \begin{equation}
        m_i\frac{dv_i}{dt} = m_i(\frac{v_i^p-v_i}{\tau_i}) + F_{int}
    \end{equation}
    where $m_i$ denotes the mass of the individual, $v_i$ indicates its actual velocity which varies given the presence of obstacles in the scene, $\tau_i$ is a relaxing parameter, $F_{int}$ indicates the interaction force encountered by the individual defined as the sum of attraction and repulsive forces, and $v_i^p$ is the desired velocity of the individual. Fig. \ref{fig:fig11} visualizes of the forces and velocities in SFM.

    Generalized SFM has been adopted as the basic model in many studies of crowd behavior analysis \cite{25,105,106}. Furthermore, the calibrated agent-based framework proposed in \cite{still2000crowd} could also accurately model a number of observed crowd phenomena.

\subsection{Crowd Model in Video Analysis}
    There has been a series of attempts to incorporate the research findings of crowd simulation to automatic crowded scenes analysis~\cite{andrade2005simulation}. Several physics-inspired crowd models have been utilized for the purposes of recognition and classification \cite{1,2,4,17,25}.

    To analyze at the macroscopic level, usually holistic properties of the scene are modeled.
    Assuming that high density crowd behaves like a complex dynamic system, many dynamical crowd evolution models have been proposed \cite{1,16,17,allain2010crowd}. The concepts of motion field and dynamical potential borrowed from fluid dynamics community were utilized\cite{ferziger1996computational}. The motion field is a rich dynamical descriptor of the flow which can be related to the velocity of flow, while the potential accounts for several physical quantities such as the density or the pressure in the flow. Although people do not always follow the laws of physics, they have choices in their direction, have no conservation of momentum and can stop and start at willing \cite{still2000crowd}, the coupling of crowd dynamics and real data has exhibited promising results in crowd video analysis and opened a rich area of research.

    At the microscopic level, agent-based models have also been popular in video analysis. They analyze the stimuli, or driven factors of crowd behavior, based on the assumption that crowd behavior originates from the interaction of its elementary individuals. Mehran \emph{et al.} \cite{25} and Zhao \emph{et al.} \cite{106} applied the SFM to detect the abnormal events of the crowd. Zhou \emph{et al.} \cite{50, 54} used dynamic pedestrian-agent model to learn the collective behavior patterns of pedestrians in crowded scenes.

    It is noted that cues from the two levels could be jointly used. For example, \cite{19, 21} employ the global motion information to improve tracking individuals in a crowd scene. Also the microscopic information of individual movements can be used as basic units in the holistic scene models.

    Besides, visual feature extraction, object tracking, learning, and other related algorithms from vision area also play important roles in crowded scene analysis. They will be given brief introductions in method descriptions.

\section{Motion Representation in Crowded Scenes} \label{sec:feature}
    As a crucial basis, an appropriate feature representation can benefit the subsequent tasks.
    Motion information representation is the basis for crowded scene analysis. Although other types of visual features such as scene structure, geometrical information and viewing direction could also be helpful, motion features are dominant in our interested area.

    According to the representation level, previous crowd motion features can be categorized into three categories: flow-based features, local spatio-temporal features and trajectory/tracklet.
    Flow-based features are extracted densely on the pixel level. Local spatio-temporal features represent the scene based on local information from 2D patches or 3D cubes. On a higher level, trajectory/tracklet tries to compute the individual tracks, as the basic features for motion representation.
    These feature representations have been used in various tasks as motion pattern segmentation, crowd behavior recognition and crowd anomaly detection.

\subsection{Flow-Based Features}
    In the context of high density crowded scenes, tracking a person or an object is always a difficult task, and sometimes unfeasible. Fortunately, when we look at crowd, we care about what is happening, not who is doing it. The specific actions of individual pedestrians may appear relatively random, but the overall look of the crowd can still be convincing \cite{leggett2004real}. For that reason, several optical-flow alike features have been presented in recent years \cite{1,16,17,25,60,61,131}, these methods avoid tracking from the macroscopic level and have achieved some success in addressing complex crowd flows in the scenes.
\begin{figure}[htbp]
        \centering
        \includegraphics[width = \linewidth]{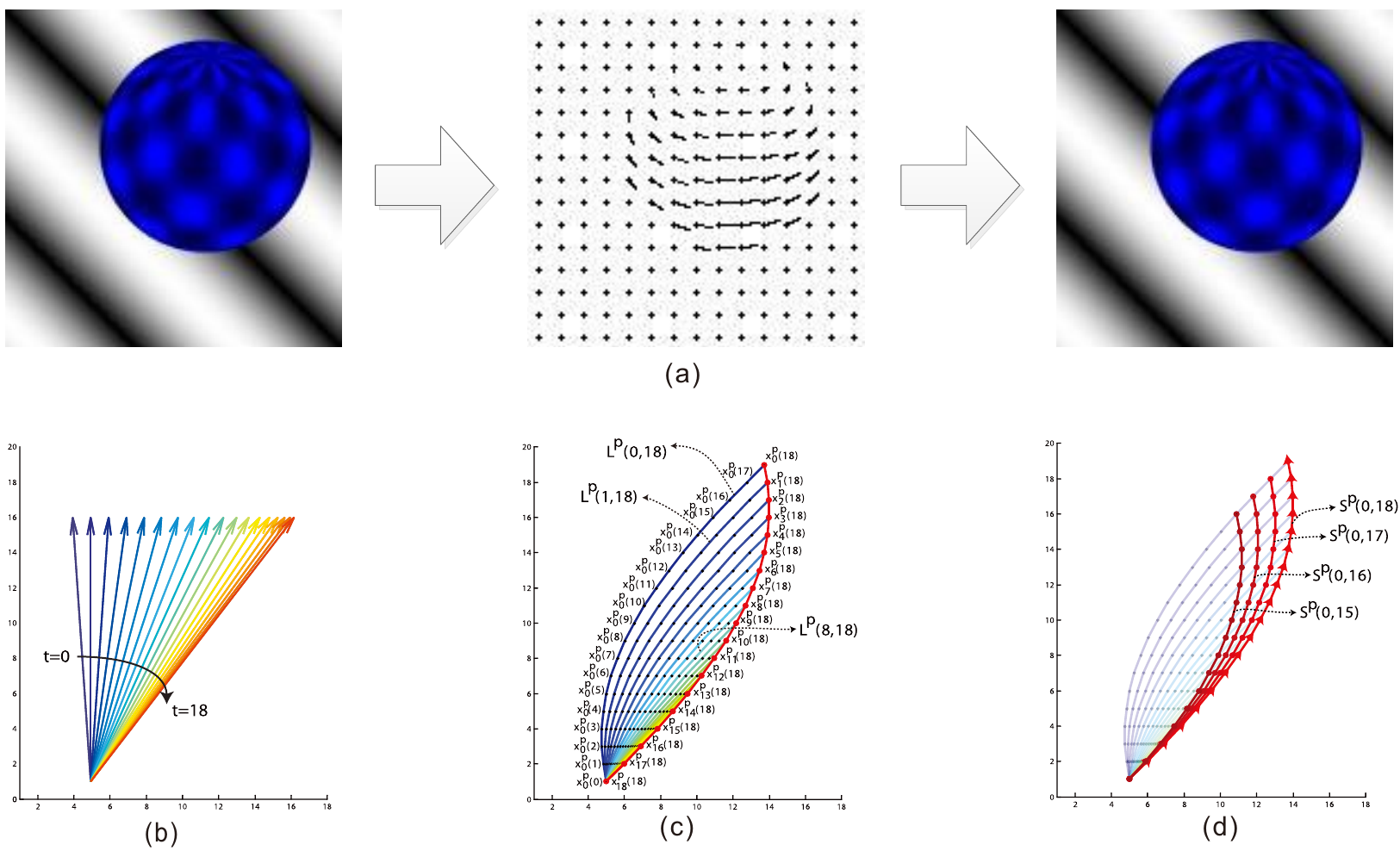}
        \caption{(a) An visualization of the optical flow feature: the rotating sphere generates the optical flow field shown in the middle. (b) (c) (d) compare the change of the feature vectors in time period \emph{t} = 0 to \emph{t} = 18 of optical flow, particle flow, and streak flow, respectively. (b) (c) (d) are originally shown in \cite{17}.}
        \label{fig:fig9}
\end{figure}
\subsubsection{Optical Flow}
    Optical flow is to compute pixel-wise instantaneous motion between consecutive frames \cite{141}. Optical flow is robust to multiple and simultaneous camera and object motions, and it is widely used in crowd motion detection and segmentation \cite{44,46,47,51,60,61}. However, optical flow does not capture long-range temporal dependencies, and can not represent spatial and temporal properties of a flow. These properties can be useful for many applications.

\subsubsection{Particle Flow}
    Recently, based on the Lagrangian framework of fluid dynamics \cite{74}, a notion of particle flow was introduced in computer vision \cite{1,16,25}. Particle flow is computed by moving a grid of particles with the optical flow through numerical integration, providing trajectories that relate a particles initial position to its position at a later time. Impressive results employing particle flow have been demonstrated on crowd segmentation \cite{16} and abnormal crowd behavior detection \cite{1,25}. However, in particle flow the spatial changes are ignored, and time delay is significant.

\subsubsection{Streak Flow}
    In order to achieve an accurate representation of the flow from crowd motion, Mehran \emph{et al.} \cite{17} introduced the notion of streakline to compute the motion field for crowd video scene analysis, referred to as streak flow. They also provided the comparison between optical flow, particle flow and streak flow with discussion. Streaklines are well known in flow visualization \cite{van2002image} and fluid mechanics \cite{ferziger1996computational} as a tool for measurement and analysis of the flow. It encapsulates motion information of the flow for a period of time. This resembles particle flow where the advection of a grid of particles provides information for segmenting the crowd motion. Streak flow exhibits changes in the flow faster than particle flow, and therefore captures crowd motions better in a dynamically changing flow.

    Fig. \ref{fig:fig9} gives an example of the optical flow feature. It also shows a comparison of optical flow, particle flow and streak flow using a locally uniform flow field changing over time.

\subsection{Local Spatio-Temporal Features}
    Some extremely crowded scenes, though similar in density, are less structural due to the high variability of pedestrian movements. The motion within each local area may be non-uniform and generated by any number of moving objects. In such circumstances, even a fine-grain representation, such as optical flow, would not provide enough motion information. One solution is to exploit the dense local motion patterns created by the subjects, and model their spatio-temporal relationships to represent the underlying intrinsic structure they form in the video \cite{20}.

    The related methods generally consider the motion as a whole, and characterize its spatio-temporal distributions based on local 2D patches or 3D cubes, such as spatio-temporal gradients \cite{20,30}, and histogram functions~\cite{46,96}.

\subsubsection{Spatio-Temporal Gradients}
    The distribution of spatio-temporal gradients has been utilized as the base representation \cite{20,30}.
    For each pixel \emph{i} in patch \emph{I}, the spatio-temporal gradient $\nabla I_i$ is calculated as
    \begin{equation}
        \nabla I_i = [I_{i,x}\ I_{i,y}\ I_{i,t}]^T = [\frac{\partial I}{\partial x}\ \frac{\partial I}{\partial y}\ \frac{\partial I}{\partial t}]^T
    \end{equation}
    where \emph{x, y} and \emph{t} are the video's horizontal, vertical, and temporal dimensions, respectively.
    The 3D gradients of each pixel collectively represent the characteristic motion pattern within the patch.
    By capturing the steady-state motion behavior with the spatio-temporal motion pattern models, Kratz \emph{et al.} \cite{20,30} demonstrated that unusual activities can be detected as statistical deviations naturally.

\subsubsection{Motion Histogram}
\begin{figure}[htbp]
\centering
        \includegraphics[width = \linewidth]{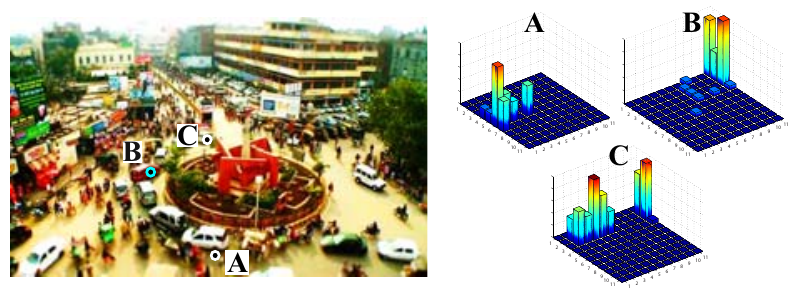}
        \caption{Motion histograms of the three pixels with different moving directions: (A) horizontal, (B) diagonal right, and (C) bimodal motion (upper left and upper right). Figure originally shown in \cite{46}.}
        \label{fig:fig12}
\end{figure}

    Motion histograms can be considered as a kind of motion information defined on local regions.
    Fig. \ref{fig:fig12} illustrates the motion histograms calculated from three sample pixels.
    In fact it originally is ill-suited for crowd motion analysis, since computing motion orientation on a motion histogram is not only time-consuming but also error-prone due to the aligning problem. Therefore, researchers have developed some more advanced features based on motion histogram \cite{46,96}. In the below brief descriptions are given.

   Jodoin \emph{et al.} \cite{46} proposed a feature called orientation distribution function (ODF). It is the probability density function of a given motion orientation. Opposed to motion histograms, ODFs have no information on the magnitude of the flow. This makes the ODF representation simpler (1D instead of 2D), and is a key advantage in computation for the upcoming motion pattern learning.

   Cong \emph{et al.} \cite{96} proposed a novel feature descriptor called multi-scale histogram of optical flow (MHOF).
   It preserves not only the motion information, but also the spatial contextual information.
   For event representation, features of all types of bases with various spatial structures are concatenated as MHOF. After estimating the motion field by optical flow, they partitioned the image into a few basic units, i.e., 2D image patches or 3D spatio-temporal cubes, then extracted MHOF from each unit.

    Overall, spatio-temporal features have shown particular promise in motion understanding due to their strong descriptive power, and therefore, they have been widely used as in various tasks such as crowd anomaly detection.

\subsection{Trajectory/Tracklet}
    Typical crowd motion is usually regular and repetitive, so that we can analyze crowd activities based on motion features extracted from trajectories of objects, 
    such as relative distance between objects, acceleration, or motion energy. 

    Comparing with the the other two types of feature representations, trajectory/tracklet is more semantic and it seems to be attractive.
    However, as mentioned previously, traditional pipeline of object detection and the subsequent tracking of those detections can hardly perform accurate object detection and tracking, as the density of the crowd increases and the scene clutter becomes severe \cite{50}.

    Considering the difficulties in obtaining complete trajectories, a motion feature called tracklet has been proposed. A tracklet is a fragment of a trajectory obtained by the tracker within a short period. Tracklets terminate when ambiguities caused by occlusions or scene clutters arise. They are more conservative and less likely to drift than long trajectories \cite{50}. In previous works \cite{65,66,67}, tracklets have been mainly used to be connected into complete trajectories for tracking or human action recognition. Recently, several tracklet based approaches for learning semantic regions and clustering trajectories \cite{29,50,54,55} were proposed. In the approaches, tracklets are often extracted from dense feature points and then a certain model is applied to enforce the spatial and temporal coherence between tracklets, to finally detect behavior patterns of pedestrians in crowded scenes.

\subsection{Summary}
   Considering the characteristics of crowded scenes, such as the congestion level, the degree of occlusion between objects, whether there is obvious motion structure, the size of field view of camera, and the resolution of a single target in the scene, we can briefly summary to which scenarios these features are suitable:
    \begin{itemize}
      \item For outdoor scenes, often with wide field of view but low resolution for each target, we mainly aim to analyze the holistic crowd motion trends. The optical flow alike features are suitable. These features can be easily combined with flow-based models, which has priority in structured crowded scene analysis.
      \item For indoor scenes, e.g., subway stations and shopping malls, the resolution of a single target is high enough and the crowd density may not be so high. The object trajectories or tracklet-based features may be good choices. These features are usually combined with agent-based models, which is applicable for analyzing activities or semantic regions in unstructured crowded scenes.
      \item Besides, when the field of view is not wide, but crowd density is high with severe occlusion, neither optical flow alike features nor tracklets will meet the requirements. Various local spatio-temporal features could be considered.
    \end{itemize}

\section{Crowd Motion Pattern Segmentation} \label{sec:segmentation}
    Motion patterns learning is important in automated visual surveillance \cite{48,49,51}. In crowded scene analysis, it is highly desirable to analyze the motion patterns and obtain some high-level interpretation.

    The term \emph{motion pattern} here refers to a spatial region of the scene that has a high degree of local similarity of the speed, as well as flow direction within the region and otherwise outside \cite{saleemi2010scene}. Motion patterns not only describe the segmentation in the spatial space, but also reflect the motion tendency in a period.
    These patterns can be joint or disjoint in the image space. They usually have a semantic level interpretation and contain sources and sinks of the paths described by the patterns.

    To analyze motion patterns in crowded scenes, various methods have been proposed. According to the principle to segment or cluster the motions, these methods can be divided into three categories: flow field model based segmentation, similarity based clustering, and probability model based clustering.
    The first category tries to simulate the image spatial segmentation based on flow field models, and therefore tends to produce spatially continuous segments.
    The later two categories utilize various well-developed clustering algorithms, usually based on local motion features, e.g., tacklets or motion video words. The resulting segments may be scattered, but they could be applicable to unstructured scenes with complex motions.

\subsection{Flow Field Model Based Segmentation}
    Among many physical-based models applied in crowd analysis, flow field models~\cite{15,16,17,46,60,61,131} are well studied in crowd motion pattern segmentation. By treating a moving crowd as a time dependent flow field consisting of regions with qualitatively different dynamics, the motion patterns emerging from spatio-temporal interactions of the participants can be reflected. Based on this kind of representations, methods such as edge-based segmentation, graph-based segmentation, watershed segmentation can be applied.

    Ali \emph{et al.} \cite{16} proposed the Lagrangian particle dynamic to segment high density crowd flows. To uncover the spatial organization of the flow field, clouds of particles generated by the crowd motion are examined. Then Lagrangian coherent structures (LCS) \cite{74} is utilized to map to the boundaries of different crowd segments. Similar to the edges for image, the LCSs for flow data can be used to segment flow regions of different dynamics. The proposed method of \cite{16} could reveal the underlying flow structures of velocity field, and it is insensitive to the scene density.
    However, slow motions might not be segmented out, and low crowd density might cause over-segmented.

    To detect typical motion patterns in crowded scenes, Hu \emph{et al.} \cite{60} constructed a directed neighborhood graph to measure the closeness of motion flow vectors, and then grouped them into motion patterns. Based on the same idea, Hu \emph{et al.} \cite{61} later invented a method to learn dominant motion patterns in videos. This is accomplished by first detecting the representative modes (sinks) of motion patterns, followed by construction of the super tracks, i.e., collective representations of the discovered motion patterns. The methods do not require complete trajectories, avoiding the problem of occlusion. But they are not applicable to unstructured scenes and the number of motion patterns needs to be predefined.

    Another approach is local-translational domain segmentation (LTDS) model proposed in Wu \emph{et al.} \cite{15}. Local crowd motion is approximated as a translational motion field, and the evolution of domain boundaries is derived from the G$\hat{a}$teaux derivative of an objective function. To represent crowd motion in an accurate and efficient way, optical flow is computed at salient locations instead of all the pixel locations. Then the problem of crowd motion partitioning is transformed to scattered motion field segmentation. This method can automatically determine the number of groups and can be applied to both medium and high density crowded scenes.

    As introduced previously in section \ref{sec:feature}, the streakline framework \cite{17} can recognize spatio-temporal flow changes more quickly than other methods. However, this framework computes the optical flow field using the conventional method. It has poor anti-interference performance, and serious deviation would be brought to the computation of the optical flow field. In \cite{131}, Wang \emph{et al.} improved the streakline framework with a highly accurate variational model \cite{141}. Different motion patterns are separated in crowded scenes by computing the similarity of streaklines and streak flows using watershed segmentation. This method finds a balance between recognition of local spatial changes and filling spatial gaps in the flow. But it must be noted that, the streak flow computation is vulnerable to disturbance, which may result in incorrect segmentation.

    Flow field model based segmentation has shown success in handling high density scenes with complex crowd flows. However, the optical flow alike features are designed to detect local changes, not to recover long-range motion patterns. Alternatively, the particle flow methods used in \cite{4,17,46} treat the crowd as a single entity, and ignore the spatial changes in coherent motion patterns. Another disadvantage of the methods in \cite{4,60,61} is that they can not handle overlapping motion patterns, since each pixel is assigned to only one motion label. It is often not the case in unstructured scenes. In addition, with the decrease in crowd density, flow field model would no longer work, and cause the video scene to be over-segmented.
\renewcommand{\arraystretch}{1.4}
\begin{table*}[tbp]
\centering
\caption{\textbf{A Brief Summary of Motion Pattern Analysis Techniques}} \label{tab:tab1}
    \begin{tabularx}{0.95\linewidth}{rYYYY}
        \toprule[2pt]
        \textbf{\small References} & \textbf{\small Datasets} & \textbf{\small Features} & \textbf{\small Applicable Scenes} & \textbf{\small Density level} \\
        \hline

        \multicolumn{5}{c}{\textbf{\small Flow Field Model Based Segmentation}}  \\
        Ali \emph{et al.} 2007 \cite{16}   & UCF          &  Particle Flow      & Structured  & High  \\
        Hu \emph{et al.} 2008 \cite{61}   & UCF          &  Optical Flow       & Structured  & High  \\
        Mehren \emph{et al.} 2010 \cite{17}     & UCF          &  Streak Flow        & Structured  & High  \\
        Wu \emph{et al.} 2012  \cite{15}   & UCF / UCSD   &  Optical Flow       & Structured  & High / Medium / Low \\
        He \emph{et al.} 2012 \cite{78}    & UCF          &  Optical Flow       & Structured  & Medium / Low \\

        \hline

        \multicolumn{5}{c}{\textbf{\small Similarity Based Clustering}} \\
        Cheriyadat \emph{et al.} 2008 \cite{13}   & \cite{13}        & Optical Flow  & Structured       & Low     \\
        Zhou \emph{et al.} 2012 \cite{77}         & UCF / \cite{77}  & Tracklet      & Structured / Unstructured   & High / Medium \\
        Wang \emph{et al.} 2013 \cite{29}         & UCF              & Tracklet      & Structured / Unstructured   & High / Medium \\
        Jodoin \emph{et al.} 2013 \cite{46}  & MIT Traffic / UCF / \cite{46} & Motion Histogram  & Structured / Unstructured &  High / Medium / Low \\

        \hline

        \multicolumn{5}{c}{\textbf{\small Probability Model Based Clustering}} \\
        Yang \emph{et al.} 2009 \cite{44}         & MIT Traffic                  & Optical Flow      & Structured      & Low         \\
        Saleemi \emph{et al.} 2010  \cite{saleemi2010scene}       & MIT Traffic  & Optical Flow      & Structured      & Medium / Low     \\
        Song \emph{et al.} 2011  \cite{6}         & MIT Traffic                          & Optical Flow      & Structured / Unstructured  & Low         \\
        Zhou \emph{et al.} 2011 \cite{50}         & CUHK                                 & Tracklet          & Structured / Unstructured  & Medium         \\
        Fu \emph{et al.} 2012   \cite{11}         & UCF / QMUL                           & Motion Histogram  & Structured / Unstructured  & High / Medium / Low \\

        \bottomrule[2pt]
    \end{tabularx}
\end{table*}

\subsection{Similarity Based Clustering}
    In this kind of methods, motion pattern segmentation is treated as a clustering problem: once motion features are detected and extracted, they are grouped into similar categories through some similarity measurements. Then, the semantic regions are estimated from the spatial extents of trajectory/tracklet clusters. The detailed descriptions are as follows.

    Cheriyadat \emph{et al.} \cite{13} used a distance measure for feature trajectories based on longest common subsequence (LCSS) \cite{7}. The method begins with independently tracking low-level features using optical flow, and then clusters these tracks into smooth dominant motions. It has a great advantage of speed in measuring the similarity between all pairs of tracks. But the parameters for clustering need to be fine-tuned for different situations. In addition, feature points tracking may suffer from noise.

    Based on tracklets, Zhao \emph{et al.} \cite{134} used a manifold learning method to infer the local geometric structures in image space, and to infer the motion patterns in videos. They embedded tracklet points into (\emph{x, y, $\theta$}) space, where (\emph{x, y}) stand for the image space and $\theta$ represents motion direction. In this space, points automatically form intrinsic manifold structures each corresponding to a motion pattern.

    Also based on tracklets, Zhou \emph{et al.} \cite{77} proposed a general technique of detecting coherent motion patterns in noisy time-series data, named coherent filtering. When applying this technique to coherent motion detection in the crowd, tracklets are firstly extracted by Kanade-Lucas-Tomasi (KLT) \cite{72} keypoint tracker. Then a similarity measure called coherent neighbor invariance is used to characterize these tracklets and cluster them into different motion patterns.
    An approach similar to \cite{77} was proposed in Wang \emph{et al.} \cite{29}, to analyze motion patterns from the tracklets in dynamical crowded scenes. Tracklets are collected by tracking dense feature points from the video of crowded scenes, and motion patterns are then learned by clustering the tracklets.

    In Jodoin \emph{et al.} \cite{46}, a method of meta-tracking has been proposed to extract dominant motion patterns and the main entry/exit areas from a surveillance video. This method relies on pixel-based orientation distributed functions (ODFs), which summarize the directions of the flows at each point of the scene. Once all pixels have been assigned to ODFs, particle trajectories are computed through an iterative algorithm. They are called ``meta-tracks''. Finally, based on a hierarchical clustering method, nearest meta-tracks are merged together to form the motion patterns.

    Compared with trajectories, local motion feature is insensitive to scene clutter and tracking errors. Through clustering or linking process, the tracklets or optical flow with common features can be properly grouped, resulting in different semantic regions. Another advantage is that local feature clustering can be used for both structured and unstructured crowded scenes, since even mutually overlapping local motion features can be well separated in the learning process.

\subsection{Probability Model Based Clustering}
    Being widely utilized in visual clustering, probability Bayesian models can also be adopted here for crowd motion pattern segmentation. Low-level motion features to be clustered are fitted with the designed models.
    Popular models in vision area, such as Gaussian mixture model (GMM), random field topic (RFT), and latent Dirichlet allocation (LDA) have been applied.
    In contrast to the simple averaging optical flow methods, the use of an probability model allows for long-term analysis of a scene. Moreover, it can capture both the overlapping behaviors at any given location in a scene and the spatial dependencies between behaviors. Finally, the statistical model can incorporate a priori knowledge on where, when and what types of activities occur.

    Yang \emph{et al.} \cite{44} proposed a novel method to automatically discover key motion patterns in a scene by observing the scene for an extended period. Firstly low-level motion features are extracted through computing optical flow. These motion features are then quantized into video words based on their direction and location. Next, some video words are screened out based on the entropy over all clips for a given word. The key motion patterns are discovered using diffusion maps embedding \cite{69} and clustering.

    For the same purpose, Saleemi \emph{et al.} \cite{saleemi2010scene} introduced a statistical model for motion patterns representation based on raw optical flow. The method is based on hierarchical problem-specific learning. GMM is exploited as a co-occurrence free measure of spatio-temporal proximity and flow similarity between features. Finally, a pixel-level representation of motion patterns is proposed by deriving conditional expectation of optical flow.

    The RFT model has been applied in semantic region analysis in crowded scenes in Zhou \emph{et al.} \cite{50, 54}, based on the motions of objects. In the approach, a tracklet is treated as a document, and observations (points) on tracklets are quantized into words according to a codebook based on their locations and velocity directions. In addition, Markov Random Fields (MRF) is used as a prior to enforce the spatial and temporal coherence between tracklets during the learning process. The MRF model encourages tracklets spatially and temporally close to have similar distributions over semantic regions. Each semantic region has its preferred source and sink. Therefore, activities observed in the same semantic region have similar semantic interpretations.

    The LDA model has also been adopted \cite{6}. Assuming that motion patterns involved in a complex dynamic scene usually have a hierarchical nature, a two-level motion pattern mining approach has been proposed. At the first level, single-agent motion patterns are modeled as distributions over pixel-based features. At the second level, interaction patterns are modeled as distributions over single-agent motion patterns. Then, LDA is applied to discover both single-agent motion patterns and interaction patterns in the video.

    Moreover, Fu \emph{et al.} \cite{11} extracted optical flow features from each pair of consecutive frames, and quantized them into discrete visual words. The video is represented by a word-document hierarchical topic model through a generative process, and an improved sparse topical coding approach is used for model learning.

    An advantage of probability models is that it can provide much more compact representations than directly clustering high dimensional motion feature vectors computed from video clips. Furthermore, it models the spatio-temporal inter-relationships among different events at the global scene level, which can facilitate the crowd behavior understanding.

\subsection{Experiments}
    To give a pilot evaluation of crowd motion pattern segmentation methods, we test five representative methods on six videos standing for different challenges. These videos were taken from the UCF datasets and \cite{46}, and their length ranges from 100 frames to 5000 frames. Some have a small number of moving objects (\emph{Pedestrians, Crosswalk, Roundabout}) while others are highly crowded (\emph{Marathon, Mecca, Pilgrims}); some have simple layout (\emph{Marathon, Mecca, Crosswalk}) while others are complex (\emph{Pedestrians, Pilgrims, Roundabout}); some have well structured dynamics (\emph{Marathon, Mecca}) while others present fairly unstructured scenes(\emph{Pedestrians}) or semi-structured scenes (\emph{Pilgrims, Crosswalk, Roundabout}).

    \begin{figure*}[tbp]
        \centering
        \includegraphics[width = 0.8\linewidth]{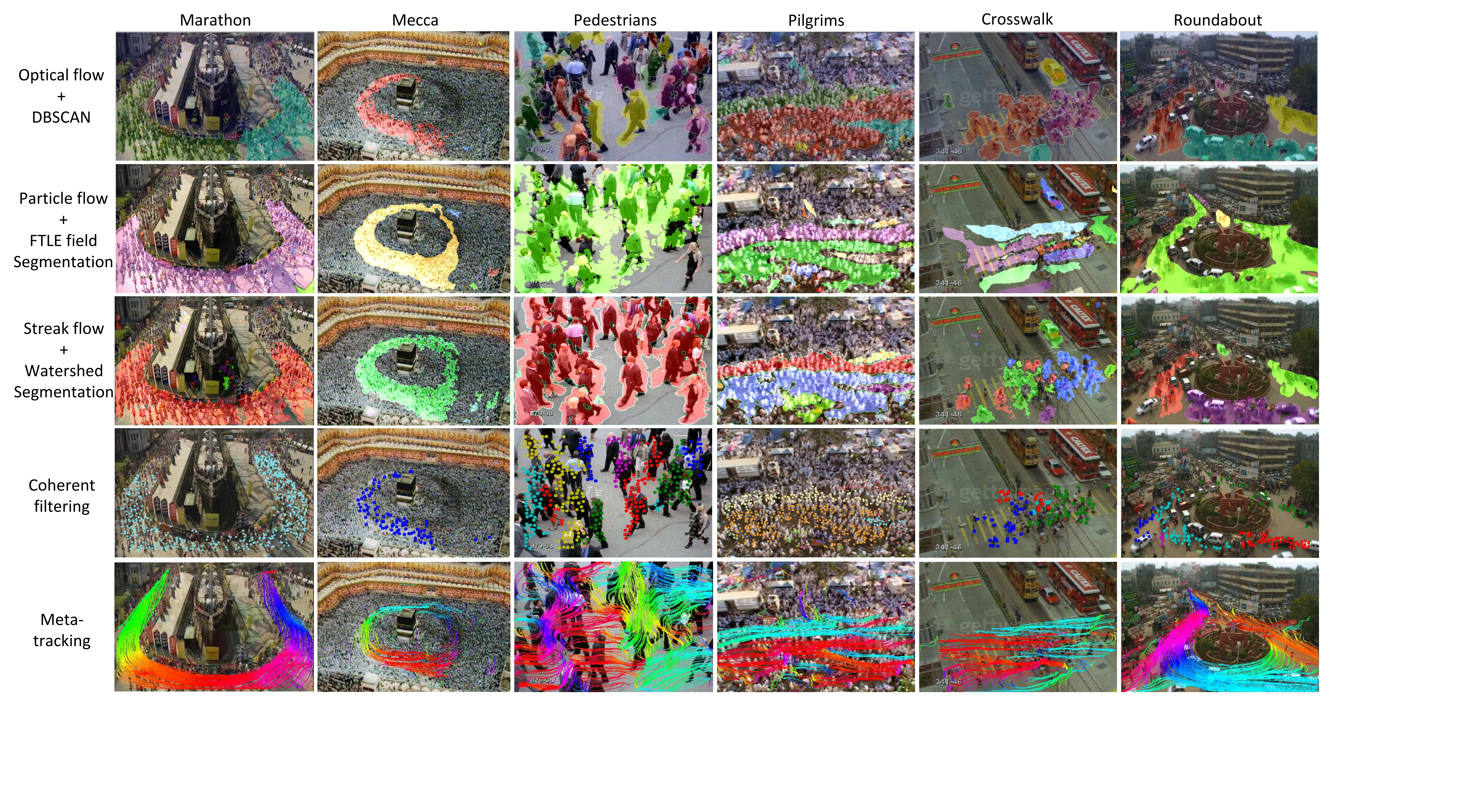}
        \caption{Motion segmentation results from the selected five methods. The first row is produced by optical flow with DBSCAN \cite{78}; the second row is produced by particle flow and FTLE field segmentation \cite{16}; the third row is produced by streak flow and watershed segmentation \cite{17}; the fourth row is produced by coherent-filtering \cite{77}; and the fifth row is produced by meta-tracking \cite{46}. For figures in the first to fourth rows, different colors represent different motion patterns. For figures in the fifth row, both color and line continuity distinguish different motion patterns. (Best viewed in color)}
        \label{fig:fig6}
    \end{figure*}

    \begin{figure*}[tbp]
        \centering
        \includegraphics[width = 0.8\linewidth]{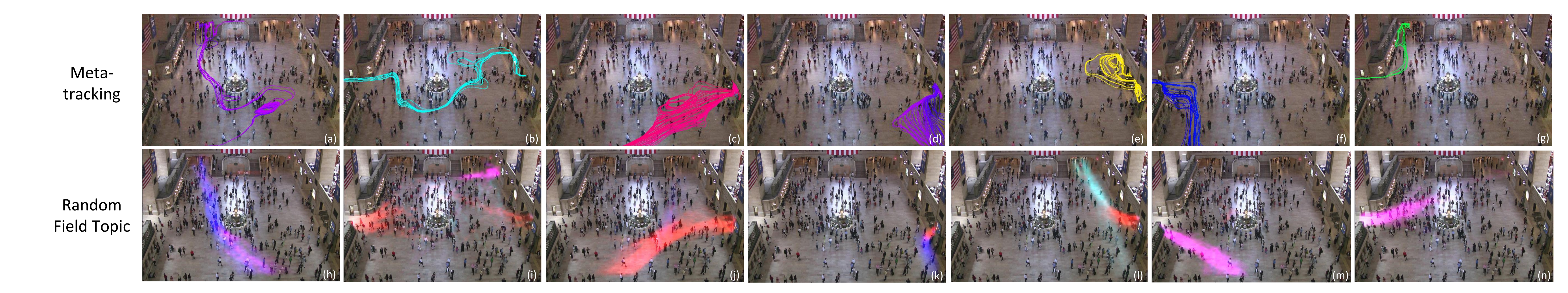}
        \caption{Part of source/sink seeking results from meta-tracking \cite{46} and RFT model \cite{50}. Different colors in the columns represent different motion patterns, each indicating a regular path of pedestrians. (Best viewed in color)}
        \label{fig:fig8}
    \end{figure*}

    Fig. \ref{fig:fig6} gives the motion segmentation results of the selected five methods on six videos. It can be found that these methods produce similar results on sequences of well structured (\emph{Marathon, Mecca}) and semi-structured scenes (\emph{Pilgrims, Crosswalk, Roundabout}), but different results on unstructured scenes (\emph{Pedestrian}).
    For structured scenes, all the methods well segment the areas with different motion characteristics, and each resulted region looks continuous and unified; for unstructured scenes, though the segmented patterns by particle flow with FTLE field and streak flow with watershed are still unified, the truth is that individuals in the resulted region move differently; the optical flow with DBSCAN (density-based spatial clustering of applications with noise) obtains poor results; the coherent-filtering produces quite scattered pieces, reflecting motion patterns at a certain point in time;
    the segmentation results of meta-tracking look quite clutter, but they mix several motion patterns together. The meta-tracking can well handle the unstructured scenes for its representative ability for multiple motion patterns.

    An experiment on video clips of 100 frames length shows that the average execution times of DBSCAN \cite{78}, FTLE \cite{16}, watershed \cite{17}, meta-tracking \cite{46} and coherent-filtering \cite{77} are around 6.3 seconds, 22.4 seconds, 5.6 seconds, 9.3 seconds and 0.4 seconds, respectively (CPU: i7-3770, 3.4GHz; memory: 8G). Here we do not consider the computation of motion features extraction. The difference in computation times lies in the fact that, FTLE and meta-tracking require motion information across the whole video sequence to generate only the motion pattern map, while DBSCAN, watershed and coherent-filtering can generate a motion pattern map with just two adjacent video frames in each iteration, which could be more efficient.

    To evaluate the motion pattern segmentation results quantitatively, we manually label the detected motion regions. The numbers of true and false detections from different methods for three representative videos (\emph{Mecca} for highly crowded structured scene, \emph{Roundabout} for semi-structured scene, and \emph{Pedestrian} for unstructured scene) are given in Table \ref{tab:tab6}. Moreover, due to the noise in motion clusters, the motion detection number cannot fully reflect the performance. We use \emph{completeness} as an additional measurement of the segmentation accuracy. Here, completeness is the ratio of the detected motion patterns area to the ground-truth area. The results are given in Table \ref{tab:tab6}, for comparison.

    Motion pattern segmentation can also be used for source/sink seeking and path planing. We also conduct an experiment on the New York grand central station video from CHUK dataset. The test video is 30 minutes long and contains an unstructured crowded scene. Since flow field based models and coherent-filtering can not handle the extremely stochastic crowded scenes, we only evaluate two methods in this experiment: meta-tracking \cite {46} and RFT model \cite{50}. From the experimental results shown in Fig. \ref{fig:fig8}, we can clearly see that some regular paths are extracted from mixed crowd motion patterns. Overall, results produced by the two methods are similar. The results from RFT look better, since the model can incorporate a prior knowledge from annotation (e.g. the number and positions of sources/sinks) in the training process.
    The meta-tracking method finds the sources/sinks via clustering, performing the whole process automatically without human intervention.

    The quantitative evaluation results are shown in Table \ref{tab:tab7}. The number of true and false sources/sinks are used. Here, a true detection means that the algorithm finds an entry/exit point accordance with the manually labeled ground-truth; while a false detection means that the algorithm finds a wrong entry/exit point. Besides, to evaluate the generated paths, we first cluster each detected path shown in Fig. \ref{fig:fig8} into a smooth dominant trajectory using the method proposed in \cite{13}, then we compute the similarity between the dominant trajectory and the labeled path. A distance measure of longest common subsequences (LCSS) \cite{vlachos2002discovering} is used. The LCSS distance between path $F_i$ and $F_j$ is defined as
    \begin{equation}
         D_{LCSS}(F_i, F_j) = \frac{LCSS(F_i, F_j)}{min(T_i, T_j)}
    \end{equation}
    where $T_i, T_j$ are the lengths of $F_i, F_j$, respectively. $LCSS(F_i, F_j)$ specifies the number of matching points between two trajectories. A good algorithm should result in high similarity values. In Table \ref{tab:tab7}, we only consider the true detected paths and compute the average LCSS distance over them.
\renewcommand{\arraystretch}{1.4}
\begin{table}[tbp]
\centering
\caption{\textbf{Motion Pattern Segmentation results}} \label{tab:tab6}
\begin{tabularx}{\linewidth}{p{1cm}ccccc}
    \toprule[1pt]
    \small{Scene} & \small{Method}     & \small{Labels}    &  \small{True}  &  \small{False}  &  \small{Complete-}     \\
                   &                    & \small{Number}     &  \small{Detections}       &  \small{Detections}    &  \small{ness}       \\
    \hline
    \multirow{5}*{Mecca}
                           & \cite{78}  & 2 &  0.67 & 1.0 &  0.65  \\
                           & \cite{16}  & 2 &  0.5  & 1.0 &  0.96  \\
                           & \cite{17}  & 2 &  0.4  & 1.0 &  0.97  \\
                           & \cite{77}  & 2 &  1.0  & 1.0 &  0.63  \\
                           & \cite{46}  & 2 &  0.67 & 1.0 &  0.92  \\
    \hline
    \multirow{5}*{Roundabout}
                           & \cite{78}  & 7  &  1.0    & 0.57   & 0.74 \\
                           & \cite{16}  & 7  &  1.0    & 0.43   & 0.94 \\
                           & \cite{17}  & 7  &  1.0    & 0.43   & 0.74 \\
                           & \cite{77}  & 7  &  0.8    & 0.57   & 0.83 \\
                           & \cite{46}  & 7  &  1.0    & 0.72   & 0.96 \\
    \hline
    \multirow{5}*{Pedestrian}
                           & \cite{78}  & 8  &  0.44   & 0.5   & 0.68 \\
                           & \cite{16}  & 8  &  1.0    & 0.125 & 0.75 \\
                           & \cite{17}  & 8  &  1.0    & 0.125 & 0.72 \\
                           & \cite{77}  & 8  &  0.5    & 0.625 & 0.70 \\
                           & \cite{46}  & 8  &  0.6    & 0.75  & 0.83 \\
    \bottomrule[1pt]
\end{tabularx}
\end{table}

\begin{table}[tbp]
\centering
\caption{\textbf{Source/Sink Seeking and Path Planing Results}} \label{tab:tab7}
\begin{tabularx}{0.48\textwidth}{p{2cm}YYYYY}
    \toprule[1pt]
    \small{Methods} & \small{Labels}  & \small{True}  & \small{False}  & \small{LCSS} \\
                    & \small{Number}  & \small{Detections} & \small{Detections}  & \small{Distance}  \\
    \hline
    Meta-tracking     & 16  & 22   &  7  & 0.73  \\
    RFT model         & 16  & 16   &  1  & 0.91 \\
    \bottomrule[1pt]
\end{tabularx}
\begin{tablenotes}
        \footnotesize
        \item Note different true detection results produced by meta-tracking \cite{46} may correspond to one ground-truth label. RFT model is referred to \cite{50}.
\end{tablenotes}
\end{table}


\subsection{Summary}
    In general, flow field model based segmentation and similarity based clustering require little human intervention. Thus motion segmentation can be performed in an unsupervised way, and it is convenient in many video analysis applications.
    Table \ref{tab:tab1} summarizes the reviewed studies on crowd motion pattern segmentation, providing the information of test dataset, applicable scene and crowd density level in the experimental settings of each method.

    Flow field model based methods are the most studied in motion pattern segmentation. Flow field can well simulate the crowd motions, considering the individuals as particles. Similarity based clustering methods are becoming more and more popular, because in high density crowds, the local motion features such as tracklets can be obtained more easily than the complete trajectories, and they show to be more discriminative than local optical flows. Recently, probability models, especially the topic models borrowed from language processing, have been applied to capture spatial and temporal dependency.
    Motion patterns in crowded scenes can be interpreted hierarchically. Individual movements constitute small group motions and they further form large motion patterns. The probability topic model may be a good choice due to its capacity to discover semantic regions and explore more details within the motion patterns.

    The learned motion patterns can be used in a range of applications including path or source/sink seeking in crowded scenes, as we have shown in experiments. Besides, several tracking algorithms \cite{18,19,20,21,22} also have interests in how to learn scene-specific motion patterns to improve the effect of tracking.

\section{Crowd Behavior Recognition} \label{sec:recognition}
    Crowd behavior analysis has been an active research topic in simulation and graphics fields where the main goal is to create realistic crowd motions \cite{25}.
    Relatively little effort has been spent on reliable classification and understanding of human activities in real-world crowded scenes. 

    In general, approaches for crowd behavior analysis can be divided to ``holistic'' and ``object-based''. The former treats the crowd as a single entity, which may be suitable to structured scenes of medium or high density \cite{36}, while the later treats the crowd as a collection of individuals.
    In holistic approaches, crowd dynamics models are usually adopted to judge the behaviors on the whole. But local behaviors in unstructured scenes can not be handled. Object-based approaches infer both the behaviors and their associated individuals.

\subsection{Holistic Approach}
    In highly crowded surveillance scenes, moving objects in the sensor range appear small or even unresolved. Very few features can be detected and extracted from an individual object. In such situations, understanding crowd behaviors without knowing the actions of individuals is often advantageous \cite{27}.

    Assuming the crowd fluid is incompressible and irrotational, Mehran \emph{et al.} \cite{17} proposed a concept of potential functions, which consists of two parts: \emph{stream function} and \emph{velocity function}. The former provides information regarding the steady and non-divergent part of the flow, whereas the later contains information regarding the local changes in the non-curling motions. With this perspective, the potential function field is capable of discriminating lanes and divergent/convergent regions in different scenes.

    To detect major motion patterns and crowd events, Benabbas \emph{et al.} \cite{51} first clustered low-level motion features to learn the direction and magnitude models of crowds, and then used a region-based segmentation algorithm to generate different motion patterns. After that, crowd events such as merge, split, walk, run, local dispersion, and evacuation were detected by analyzing the instantaneous optical flow vectors and comparing with the learned models.

    Later in Solmaz \emph{et al.} \cite{27}, crowd behaviors including bottlenecks, fountainheads, lanes, arches, and blockings can be recognized. In this framework, a scene is represented by a grid of particles initializing a dynamical system defined by the optical flow. Time integration of the dynamical system provides particle trajectories that represent the motion in the scene. These trajectories are used to locate regions of interest in the scene. Behavior classification is obtained according to the Jacobian matrix based on linear approximation of the dynamical system. The eigenvalues are used to determine the dynamic stability of points in the flow and each type of stability corresponds to one of the five crowd behaviors.

    Recently, a similar spatio-temporal viscous fluid (STVF) field was adopted in Su \emph{et al.} \cite{47}, to model crowd motion patterns by exploring both appearance of crowd behaviors and interaction among pedestrians. In their approach, large-scale crowd behaviors are recognized based on the characteristics of the fluid field. First, a spatio-temporal variation matrix is proposed to measure the local fluctuation for specific pixels. Then, the force among pedestrians are modeled with shear force in the spatio-temporal variation fluid field. Finally, a codebook is constructed by clustering neighboring pixels with similar spatio-temporal features, and crowd behaviors are recognized using the LDA model.

\renewcommand{\arraystretch}{1.4}
\begin{table*}[htbp]
\centering
\caption{\textbf{Comparison of crowd behavior recognition techniques}}\label{tab:tab2}
\begin{threeparttable}
    \begin{tabularx}{0.9\linewidth}{rrYrrr}
        \toprule[2pt]
        \multirow{2}*{\textbf{\small Reference}}  & \multirow{2}*{\textbf{\small Dataset}} & \multirow{2}*{\textbf{\small Method}} & \multicolumn{3}{c}{\textbf{\small Performance}} \\
        \cline{4-6}                               &                                        &                                      & \textbf{\small AUC} & \textbf{\small ACC} & \textbf{\small DR} \\

        \hline
        \multicolumn{6}{c}{\textbf{\small Holistic Approach}} \\
        Mehran 2010 \cite{17}   & UMN / \cite{17}      & Streak flow field + Potential field          & $\approx$ 0.848   & N/A                & $\approx$ 97.0\% \\
        Benabbas 2011 \cite{51} & PETS2009 / \cite{51} & Optical flow field + Block clustering        & N/A               & $\approx$ 77.0\%   & N/A   \\
        Solmaz 2012 \cite{27}   & PETS2009 / \cite{27} & Eigenvalue analysis + Stability analysis     & $\approx$ 0.825   & $\approx$ 71.2\%   & $\approx$ 85.3\%  \\
        Su 2013 \cite{47}       & PETS2009 / UMN       & STVF + LDA                                   & 0.873	          & 87.9\%             & N/A    \\

        \hline

        \multicolumn{6}{c}{\textbf{\small Object-Based Approach}} \\
        Wang 2009 \cite{24}  & \cite{24}               & Hierarchical Bayesian Models              & N/A & 85.7\% & N/A \\
        Zhou 2012 \cite{54}  & \cite{54}               & MDA                                       & N/A & N/A    & N/A \\
        \bottomrule[2pt]
        \end{tabularx}

        \begin{tablenotes}
        \footnotesize
        \item[1] \emph{AUC: \textbf{A}rea \textbf{U}nder ROC \textbf{C}urve,}
                 \emph{ACC: \textbf{A}ccuracy, }
                 \emph{DR:  \textbf{D}etection \textbf{R}ate,}
                 \emph{N/A: \textbf{N}ot \textbf{A}vailable}
        \item[2] Note that the ACC result of \cite{24} is on the video segmentation task for different types of interactions; in \cite{54}, the average prediction error of individual behavior is shown to be $\approx$ 68 pixels/140 frames.
    \end{tablenotes}
\end{threeparttable}
\end{table*}

\subsection{Object-Based Approach}
    In unstructured crowded scenes, considering the crowd as one entity would fail to identify abnormal events that arise due to inappropriate actions of an individual. For instance, a running person in a crowd can indicate an abnormal event if the rest of the crowd are walking.
    Object-based methods may overcome this problem. Without considering the high crowd density, conventional approaches for behavior analysis are usually performed based on detection and segmentation of each individual. They suffer from the complexity to isolate individuals in the dense crowd. Addressing this, some researchers have extended this kind of approaches to highly crowd scenes, by utilizing low-level features and probability models instead of tracking a single object \cite{24, 54}.

    Wang \emph{et al.} \cite{24} used hierarchical Bayesian models to connect three elements in visual surveillance: low-level visual features, simple ``atomic'' activities, and interactions. Atomic local motions are classified into atomic activities if they are observed in certain semantic regions. The global behaviors of video clips are modeled based on the distributions of low-level visual features, and multi-agent interactions are modeled based on the distributions of atomic activities. Without labeled training data and tracking procedure, the framework fulfils many challenging visual surveillance tasks, such as segmenting motions into different activities and supporting high-level queries on activities and interactions.

    Later in \cite{54}, Zhou \emph{et al.} proposed a mixture model of dynamics pedestrian-agents (MDA) to learn the collective behavior patterns of pedestrians in crowded scenes. In the agent-based modeling, each pedestrian in the crowd is driven by a dynamic pedestrian-agent and the whole crowd is modeled as a mixture of dynamic pedestrian-agents. Once the model is learned from the training data, MDA can well infer the past behaviors, predict the future behaviors of pedestrians given their partially observed trajectories, and classify different pedestrian behaviors in the scene. However, some limitations exit, e.g., MDA assumes affine transform, and it has difficulty in representing some complex shapes.

\subsection{Summary}
    The principle to classify crowd behavior recognition methods depends on the perspective from which we observe the crowd: a single entity or a bunch of independent individuals. Holistic approaches are to classify the whole streams of people to normal or abnormal, or recognize the predefined crowd behaviors. This kind of methods ignore individual difference, and consider all individuals in the crowd to have similar motion characteristics. Such a hypothesis allows us to analyze the crowd states of behavior from a systematic perspective. However, without information from object detection and tracking, a particular activity cannot be separated from other activities simultaneously occurring in the same stream. 

    In contrast, object-based approaches are able to locate typical activities and interactions in the scene, detect normal and abnormal activities, and support high-level semantic queries on activities and interactions. However, these methods can not handle dense crowded scenes, where individual objects detection does not work, and the crowd dynamics in this area appears chaotic. Under such circumstance, the spatial distribution of low-level visual features is also chaotic, and subsequent clustering procedure will not work well.

    Table \ref{tab:tab2} lists the representative crowd behavior recognition techniques, as well as their reported performances.
    A missing entry means that the quantitative result is not reported in the available literature. Besides, the crowd behaviors defined in different works are not the same. So, it is impossible to directly compare the performances of these methods, and each of the relevant studies has been conducted under different experimental conditions, using different data and different evaluation criteria.

\section{Crowd Anomaly Detection} \label{sec:anomaly}
    Anomaly detection is a key aspect of the crowded scene analysis, which has attracted much attention \cite{1,16,17,25,30,94,95,96,97,98,99}. 
    However, the problem of anomaly detection is still greatly open, and research efforts are scattered not only in approaches, but also in the interpretation of the problem, assumptions and objectives \cite{2}.

    Crowd anomaly detection methods could be learned on different supervision levels: from data with labels of both normal and abnormal behaviors; or from a corpus of unlabeled data, assuming that most parts are normal.

    Depending on the scale of interest, previous studies on anomaly detection can be categorized into two classes: global anomaly detection and local anomaly detection \cite{96}, namely ``does the scene contain an anomaly or not?" and ``where is the anomaly taking place?". 
    Detailed descriptions will be given as follows.

\subsection{Global Anomaly Detection}
    Usually, the self-organization effects occurring in crowds result in regular motion patterns. However, when abnormal events affecting public safety happen, such as fires, explosions, transportation disasters, people would escape, and it causes the crowd dynamics into a completely different state. Global anomaly detection aims to distinguish the abnormal states of crowd from normal ones. Related methodologies usually tend to detect the changes or events based on the apparent motion estimated on the whole.
    It is also important for a global anomaly detection system to not only do well in detecting the presence of anomaly in the scene, but also accurately determine the starting and end of the events, as well as the transitions between them.

    It should be noticed that holistic approaches for crowd behavior recognition mentioned in Section \ref{sec:recognition}, such as \cite{17,27,47,51}, can be applied for global crowd anomalies detection. 
    There also exist some works specifically for anomaly detection, in global style.

    In Chen \emph{et al.} \cite{108}, each isolated region is considered as a vertex and a human crowd is represented with a graph. To effectively model the topology variations, local characteristics (e.g. triangle deformations and eigenvalue-based subgraph analysis), and global features (e.g. moments) are used. They are finally combined as an indicator to detect if any anomaly of the crowd presents in the scene.

    Recently, a Bayesian framework for crowd escape behavior detection in videos was proposed \cite{wu2014bayesian}, to directly model crowd motions as non-escape and escape. Crowd motions are characterized using optical flow fields, and the associated class-conditional probability density functions are constructed based on the field attributes. Crowd escape behavior can be detected by a Bayesian formulation. Experiments demonstrated that the method is more accurate than state-of-the-art techniques in detecting crowd escape behavior. However, this method cannot be applied to high density crowded scenes for the moment, since the crowd escape behavior in this case is significantly different from that in low or medium density crowded scenes.

\subsection{Local Anomaly Detection}
    In addition to global anomaly detection, we often need to know where the anomaly event happens. To this aim, various local anomaly detection techniques have been proposed. Popular models from both crowd dynamics (e.g., flow field model, social force model, and crowd energy model) and vision area (hidden Markov model, dynamic texture, bag-of-words, sparse representation, and manifold learning) have been applied here.

    We divide the methods into two classes:
    vision-based approaches that learn model and predict anomaly purely using techniques from computer vision area, based on visual features;
    and physics-inspired approaches that incorporate related physical models for crowd dynamics representation, and achieve anomaly detection with learning methods.

\renewcommand{\arraystretch}{1.4}
\begin{table*}[htbp]
\centering
\caption{\textbf{Comparison of anomaly detection techniques}}\label{tab:tab3}
\begin{threeparttable}
    \begin{tabularx}{0.95\linewidth}{rrYccYrrrr}
        \toprule[2pt]
        \multirow{2}*{\textbf{\small Reference}} & \multirow{2}*{\textbf{\small Dataset}} & \multirow{2}*{\textbf{\small Method}} & \multicolumn{2}{c}{\textbf{\small Application} } & & \multicolumn{4}{c}{\textbf{\small Performance}} \\
        \cline{4-5} \cline{7-10} & & & \textbf{\small Global} & \textbf{\small Local} &  & \textbf{\small AUC} & \textbf{\small ACC} & \textbf{\small DR} & \textbf{\small ERR} \\

        \hline
        \multicolumn{10}{c}{\textbf{\small Vision-Based Approach}} \\
        Mahadevan 2010 \cite{94}    & UCSD             & MDT                             &         & $\surd$ & & $\approx$ 0.735     & 75\%           & N/A        & 25\%   \\
        Benabbas 2011 \cite{51}     & PETS2009 / CUHK  & Optical flow + Block clustering & $\surd$ &         & &  N/A                & $\approx$ 77\% & N/A        & N/A    \\
        Wang 2012 \cite{95}         & UMN		       & HSFT + HMMs                     &         & $\surd$ & & $\approx$ 0.900     & N/A            & 85\%       & N/A    \\
        Cong 2012 \cite{96}         & UMN / UCSD       & SRC + LSDS                      & $\surd$ & $\surd$ & & G: 0.980            & G: N/A    & G: N/A     & 20\%   \\
                                    &                  &                                 &         &         & & L: 0.478            & L: 46\%   & L: 46\%    &        \\
        Cong 2013  \cite{99}        & UCSD		       & STMC + DPG                      &         & $\surd$ & & 0.868               & N/A       & N/A        & 23.9\% \\
        Roshtkhari 2013 \cite{97}   & UCSD             & STC + BOW                       &         & $\surd$ & & $\approx$ 0.879     & N/A       & 95.8\%     & 20.5\% \\
        Thida 2013 \cite{98}        & UMN / UCSD       & SLE                             &         & $\surd$ & & $\approx$ 0.977     & N/A       & N/A        & 17.8\% \\
        Li 2013 \cite{100}          & UMN / \cite{100} & MDT                             &         & $\surd$ & & 0.995               & N/A            & N/A        & 3.7\%  \\	
        Chen 2013 \cite{108}        & UMN              & Eigenvalue-based graph analysis & $\surd$ &         & & N/A                 & N/A            & 91\%       & N/A    \\
        Wu 2014 \cite{wu2014bayesian} & UMN / PETS2009 & Optical flow + Bayes Classification & $\surd$ &     & & N/A                 & $\approx$ 91.57\%  & N/A    & N/A    \\

        \hline

        \multicolumn{10}{c}{\textbf{\small Physics-Inspired Approach}} \\
        Mehran 2009 \cite{25}       & UMN / \cite{25}  & SFM + LDA                       &         & $\surd$ & & 0.960               & N/A       & N/A        & N/A    \\
        Wu 2010 \cite{1}            & UMN              & Chaotic invariance              &         & $\surd$ & & 0.990               & N/A       & N/A        & N/A    \\
        Raghavendra 2011 \cite{105} & UCSD             & Particle flow + PSO-SFM         &         & $\surd$ & & 0.875               & N/A       & 52\%       & 17.0\% \\
        Zhao 2011 \cite{106}        & UMN              & Velocity-field-based SFM        &          & $\surd$ & & $\approx$ 0.940     & N/A       & N/A        & N/A    \\
        Yang 2012 \cite{101}        & UMN              & Local pressure model            & $\surd$ &         & & $\approx$ 0.975     & N/A       & N/A        & N/A    \\
        Ren 2012 \cite{104}         & UMN / UCSD       & Behavior entropy model          & $\surd$ & $\surd$ & & 0.893               & N/A       & N/A        & N/A    \\

        \bottomrule[2pt]
    \end{tabularx}

    \begin{tablenotes}
    \footnotesize
        \item[1] \emph{AUC:  \textbf{A}rea \textbf{U}nder ROC \textbf{C}urve, }
                 \emph{ACC:  \textbf{A}ccuracy, }
                 \emph{DR:   \textbf{D}etection \textbf{R}ate,}
                 \emph{ERR:  \textbf{E}qual \textbf{E}rror \textbf{R}ate,}
                 \emph{N/A:  \textbf{N}ot \textbf{A}vailable, }
                 \emph{G:    \textbf{G}lobal Detection,}
                 \emph{L:    \textbf{L}ocalization,}
                 \emph{DPG:  \textbf{D}ynamic \textbf{P}ath \textbf{G}rouping, }
                 \emph{HSFT: \textbf{H}igh-\textbf{F}requency and \textbf{S}patio-\textbf{T}emporal Feature, }
                 \emph{SLE:  \textbf{S}patio-temporal \textbf{L}aplacian \textbf{E}igenmap, }
                 \emph{STMC: \textbf{S}patio-\textbf{T}emporal \textbf{M}otion \textbf{C}ontext, }
                 \emph{STC:  \textbf{S}patio-\textbf{T}emporal \textbf{C}ompositions, }
    \end{tablenotes}
\end{threeparttable}
\end{table*}

\subsubsection{Vision-Based Approach}
    Many machine learning techniques have achieved great success in vision tasks. They have also been applied in local anomaly detection.
    These methods usually extract visual features and construct a set of clusters to represent several possible event patterns.

    \paragraph{Hidden Markov Model} The hidden Markov model (HMM) is able to take into account the inherently dynamic nature of the observed features \cite{2}. It is applicable in video event detection as well as anomaly detection.

\begin{figure}[tbp]
        \centering
        \includegraphics[width = \linewidth]{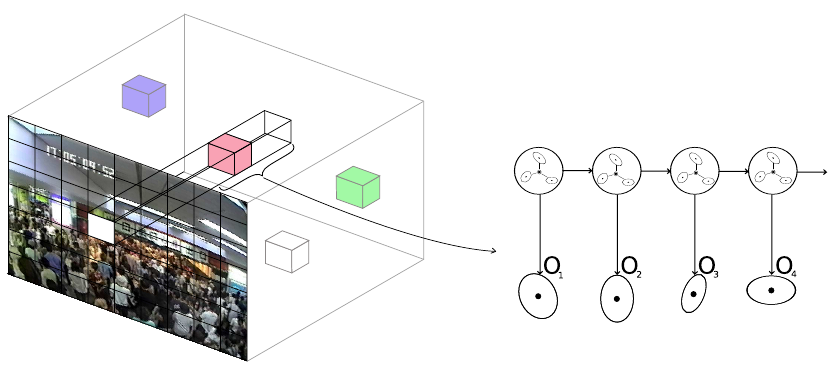}
        \caption{The temporal relationship between local motion patterns is encoded in a distribution-based HMM at each spatial location. Originally shown in \cite{30}. }
        \label{fig:fig13}
    \end{figure}

    Based on HMM, Kratz \emph{et al.} \cite{30} have presented a framework for modeling local spatio-temporal motion behaviors in extremely crowded scenes. 
    Fig. \ref{fig:fig13} illustrates a single HMM for each spatial location of observation.

    In the training phase, the temporal relationship between local motion patterns is captured via a distribution-based HMM, and the spatial relationship is modeled by a coupled HMM.
    In the testing phase, unusual events are identified as statistical deviations in video sequences of the same scene.
    The experimental results indicated that the proposed representation is suitable for analyzing extremely crowded scenes.
    However, the authors only set up one HMM for each local area, so that the method could work only for limited kinds of normal behaviors or specific crowded scenes. If we change the normal behavior type, the detection rate of the abnormal behaviors will decrease, unless the model is re-trained.

    A similar scheme has been proposed in Wang \emph{et al.} \cite{95}. In their approach, the high-frequency and spatio-temporal (HFST) information is computed by the wavelet transformation, to characterize the dynamic properties of the local region. Then, in order to detect various local abnormal crowd events, multiple HMMs are adopted, and each HMM accounts for a type of behavior.

    \paragraph{Dynamic Texture Model} The dynamic texture \cite{139} is a spatio-temporal generative model for video. It represents video sequences as observations from a linear dynamical system, and exhibits spatio-temporal stationary properties \cite{130}. Recent research works \cite{94,100} have shown that dynamic texture is more suitable for local unusual event detection in crowded scenes than optical flow.

\begin{figure}[tbp]
        \centering
        \includegraphics[width = \linewidth]{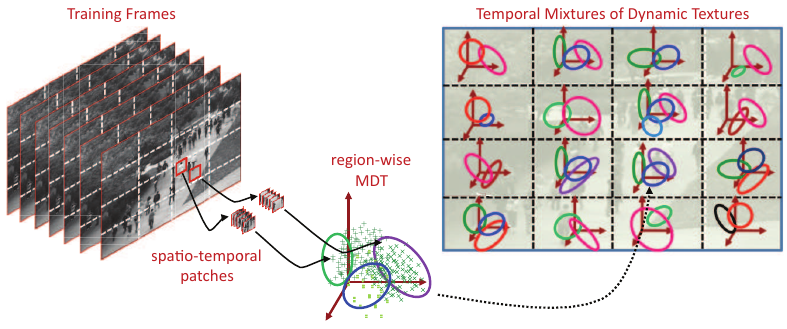}
        \caption{MDT for temporal abnormality detection. For each region of the scene, an MDT is learned. Originally shown in \cite{100}. }
        \label{fig:fig10}
    \end{figure}

    Originally proposed for motion segmentation in Chan \emph{et al.} \cite{135}, the mixture of dynamic texture (MDT) is a generative model, where a collection of video sequences are modeled as samples from a set of underlying dynamic textures.
    Fig. \ref{fig:fig10} illustrates the MDT from a video patch.
    Based on MDT, Li \emph{et al.} \cite{94,100} proposed a joint detector of temporal and spatial anomalies in crowded scenes. The proposed detector is based on a video representation that accounts for both appearance and dynamics, using a set of MDT models. The normal patterns is learned through MDT model per scene subregion, in training phase.
    A multi-scale temporal anomaly map is produced by measuring the negative log probability of each video patch under the MDT of the corresponding region. In the testing phase, subregion patches of low probability under the associated MDT are considered as anomalies.

    \paragraph{Bag-of-Words Model} One representative approach in anomaly detection is to use local spatio-temporal video volumes based bag-of-words (BOW) models. This approach usually extracts local low-level visual features, such as motion and texture, either by constructing a pixel-level background model and behavior templates, or by employing spatio-temporal video volumes \cite{97}.

    In \cite{97}, Roshtkhari \emph{et al.} extended BOW model for detecting suspicious events in videos. The method codes a video as a compact set of spatio-temporal volumes. Uncertainty is considered in the codebook construction.
    The spatio-temporal compositions of video volumes are modeled using a probabilistic framework, and anomalous events are assumed to be video presentations with very low frequency of occurrence. As a result, an observation is considered to be abnormal if it cannot be reconstructed with previous observations.

    LDA has been adopted in Wang \emph{et al.} \cite{140} based on the spatio-temporal cuboid from the video sequence with an adaptive size. To compute the similarity between two spatio-temporal cuboids with different sizes, they designed a novel data structure of codebook constructed as a set of two-level trees. LDA model is used to learn an appropriate number of topics to represent these scenarios. A new sample will be classified as anomaly if it does not belong to these topics.

    \paragraph{Sparse Representation Models} Recently, Cong \emph{et al.} \cite{96} presented a novel algorithm for abnormal event detection based on the sparse reconstruction cost (SRC) for multi-level histogram of optical flows. Given an image sequence or a collection of local spatio-temporal patches, MHOF features are calculated. Then, SRC over the normal dictionary is used to measure the normalness of the testing sample. By introducing a prior weight of each basis during sparse reconstruction, the proposed SRC is more robust than other outlier detection criteria.

    Combined with dynamic texture, Xu \emph{et al.} \cite{130} proposed a novel approach for unusual event detection via sparse reconstruction on an over-complete basis set. The dynamic texture is described by local binary patterns from three orthogonal planes (LBPTOP). In the detection process, given the basis set learned from the training procedure and the input observation, the sparse coefficients are computed and the reconstruction error is defined. The unusual events are identified as those dynamic textures with high reconstruction error.

    \paragraph{Manifold Learning Model} In \cite{98}, the manifold learning-based framework has also been applied for the detection of anomalies in a crowded scene. The spatio-temporal Lagrangian eigenmap method is employed to study the local motion structure of the scene. Besides, a pairwise graph is constructed by considering the visual context of multiple local patches in both spatial and temporal domains. Such a process embeds local motion patterns into different spatial locations where similar patterns are usually close and different patterns are far apart. This allows to cluster embedding points and to discover different motion patterns in the scene. Finally, a local probability model is used to localize the abnormal regions in the crowded scene, where the clusters with small data points or outliers in the embedded space can be considered as abnormal. \\

\subsubsection{Physics-Inspired Approach}
    Several physics-inspired models have been proposed for crowd representation, and they have also been utilized and combined with machine learning techniques for anomaly detection.
For example, continuum-based approach and agent-based approach from crowd simulation field both have been adopted for anomaly detection in crowded scenes.

    \paragraph{Flow Field Model} Usually, we need to understand how crowds evolve with time and try to find some regular patterns. So that we can know immediately where and how the motion pattern of the crowd changes. As noted in section \ref{sec:segmentation}, the work of Ali \emph{et al.} \cite{16} has shown success on motion pattern segmentation. Furthermore, they also extended their framework to anomaly detection. They constructed a finite time Lyapunov exponent (FTLE) field whose boundaries vary with the crowd changes in terms of the dynamic behavior of the flow. New Lagrangian coherent structures (LCS) \cite{74} will appear in the FTLE field exactly at those locations where the changes happen. Any change in the number of flow segments over time is regarded as an instability, and it is detected by establishing correspondences between flow segments over time.

    Wu \emph{et al.} \cite{1} proposed a method for crowd flow modeling and anomaly detection for both structured and unstructured scenes. The overall work-flow begins with particle advection based on optical flow, and particle trajectories are clustered to obtain representative trajectories for a crowd flow. Next, the chaotic dynamics of all representative trajectories are extracted and quantified using chaotic invariants. This is known as maximal Lyapunov exponent and correlation dimension in dynamic system. Probability model is learned from these chaotic feature sets. Finally, a maximum likelihood estimation criterion is adopted to identify a query video of a scene as normal or abnormal.

    Commencing with the optical flow field estimation adapted from \cite{137}, Loy \emph{et al.} \cite{129} presented a global motion saliency detection framework. The associated flow vector in the field is represented by its phase angle $ -\pi \leq \varphi_{x,y,t} \leq \pi $ and the velocity magnitude $ \gamma_{x,y,t} \geq 0 $, referred as motion signature for salient motion detection. Then the spectral residual approach \cite{138} is applied on the motion signature for motion saliency detection. This method has shown its potential in unstable region detection in extremely crowded scenes, and gave fairly similar results to \cite{16}.

    \paragraph{Social Force Model}
    The social force model (SFM) has been successfully employed in research fields as simulation and analysis of crowds.
    Mehran \emph{et al.} \cite{25} introduced a novel method to detect and localize abnormal behaviors in crowd videos using SFM \cite{helbing1995social}. For this purpose, the framework proposed by Ali \emph{et al.} \cite{16} is utilized to compute particle flows, and their interaction forces are estimated using SFM. The interaction force is then mapped into the image plane to obtain force flow for every pixel in every frame. Randomly selected spatio-temporal volumes of force flows are used to model the normal behavior of the crowd. Finally the frames are classified as normal or abnormal by using BOW. The regions of anomalies in the abnormal frames are localized using interaction forces.

    Inspired by \cite{25}, some methods based on SFM to detect abnormal crowd behaviors were later proposed. In \cite{105,136}, Raghavendra \emph{et al.} introduced the particle swarm optimization (PSO) method for optimizing the interaction force computed using SFM. The main objective of the proposed method is to drift the population of particles towards the areas of the main image motion. Such displacement is driven by the PSO fitness function, which aims at minimizing the interaction force, so as to model the most diffused and typical crowd behavior.

    A velocity-field based SFM has been proposed in Zhao \emph{et al.} \cite{106} to locate crowd behavior instability spatio-temporally. The traditional SFM defines the interaction force as a dependent variable of relative geometric position of the individuals. Differently, the proposed improved model can provide a better prediction of interactions using the collision probability in a dynamic crowd. With spatio-temporal instability analysis, we can extract video clips with potential abnormality and locate the regions of interest where the abnormalities are likely to happen.

    \paragraph{Crowd Energy Model} The crowd has its own characteristics. For example, local density and velocity are key parameters for measuring the crowd dynamics. Yang \emph{et al.} \cite{101} proposed an efficient method based on the histogram of oriented pressure (HOP) to detect crowd anomaly. SFM and local binary pattern (LBP) are adopted to calculate the pressure. Cross histogram is utilized to produce the feature vector instead of parallel merging the magnitude histogram and direction histogram. Afterwards, support vector machine and median filter are adopted to detect the anomaly.

    In \cite{102}, Xiong \emph{et al.} proposed a novel method to detect two typical abnormal activities: pedestrian gathering and running. The method is based on the potential energy \cite{103} and kinetic energy. A term called crowd distribution index (CDI) is defined to represent the dispersion, which can later determine the kinetic energy. Finally the abnormal activities are detected through threshold analysis.

    Another abnormal crowd behavior detection model using behavior entropy has been proposed in~\cite{104}. The key idea is to analyze the change of scene behavior entropy (SBE) over time, and localize abnormal behaviors according to pixels' behavior entropy distribution in image space. Experiments reveal that SBE of the frame will rise when running, dispersion, gathering or regressive walking occurs.

    This energy-based model can well denote the dispersion on different directions and locate moving information and interacting information among individuals. The model works owing to the fact that obvious differences exist between normal states and abnormal states in crowd dynamic characteristics. Usually some threshold-based methods are employed here, and the threshold usually has to be determined empirically when applied to different crowd scenes.

\subsection{Summary}
    It is usually difficult to compare different methods objectively, since anomalies are often defined in a somewhat subjective form, sometimes according to what the algorithms can detect \cite{100}. 
    We make a brief comparison of recently developed anomaly detection techniques in Table \ref{tab:tab3}. These techniques are evaluated on different datasets with different criteria. It is hard to compare them directly, and this table intends to provide a quick way to understand the solutions on the whole.

    Methods based on knowledge from physical systems for crowd representation are convenient to apply but their capacity is limited to recognize certain patterns~\cite{popoola2012video}. In order to overcome these limitations, data-driven learning-based models can be utilized or be combined to represent the events and structures of the scene.
    Among them, generative topic models seem to be promising in the interested area \cite{97,140}.
    The topic models share a fundamental idea that ``a crowded scene with its various events can be simulated as a document with mixture of topics". Characterizing unusual events by low word-topic probabilities far from existing typical topics, the topic models have the ability to automatically discover meaningful events or activities from the co-occurrences of visual words.

    In addition, recently, some techniques in video event detection, such as spatio-temporal path searching \cite{tran2013video} and background substraction \cite{tian2013selective} have been extended to crowded scenes. They can be utilized for improving crowd event detection.

\section{Crowd Video Datasets} \label{sec:dataset}
\renewcommand{\arraystretch}{1.4}
\begin{table*}[htbp]
\centering
\caption{\textbf{Datasets for Crowded Scene Analysis}} \label{tab:tab5}
\begin{tabularx}{0.95\textwidth}{rp{8cm}llY}
    \toprule[2pt]
    \textbf{\small Reference} & \multicolumn{1}{c}{\textbf{\small Description}} & \textbf{\small Size}  & \textbf{\small Label} & \textbf{\small Accessibility}  \\
    \hline
    UCF \cite{16}            & Videos of crowds, vehicle flows and other high density moving objects  & 38 videos       & Partial  & Yes  \\
    UMN \cite{111}            & Scenarios of an escape event in 3 different indoor and outdoor scenes  & 11 videos       & All      & Yes  \\
    UCSD \cite{112}           & Subset1: 34 training video clips and 36 testing video clips            & 98 video clips  & All      & Yes  \\
                              & Subset2: 16 training video clips and 12 testing video clips            &                 &          &      \\
    CUHK \cite{24}           & 1 traffic video sequence and 1 crowd video sequence                    & 2 videos        & Partial  & Yes  \\
    QMUL \cite{115}           & 3 traffic video sequence and 1 pedestrians video sequence              & 4 videos        & Partial  & Yes  \\
    PETS2009 \cite{116}       & 8 video sequences of different crowd activities with calibration data  & 8 videos        & All      & Yes  \\
    VIOLENT-FLOWS \cite{113}  & Real-world video sequences of crowd violence                           & 246 videos      & Partial  & Yes  \\
    Rodriguez's \cite{22}     & Large collection of crowd videos with 100 labeled object trajectories & 520 videos      & Partial  & No   \\
    UCF Behavior \cite{27}    & image sequences from the web videos and PETS2009          & 61 sequences       & All  & Yes  \\
    \bottomrule[2pt]
\end{tabularx}
\end{table*}

    With the development of crowded scene analysis, several crowd datasets are available now. In the following, we will list some existing benchmark crowd video datasets.
    In Table \ref{tab:tab5}, the following information is given: brief descriptions, size of each database, the labeling level and the accessibility.

    \textbf{UCF Crowd Dataset} \cite{16} This crowd dataset is collected mainly from the BBC Motion Gallery and Getty Images website, and it is publicly available. The most distinguishing feature is its variations in lighting and field of view, which can facilitate the performance evaluation of algorithms developed for crowded scenes.

    \textbf{UMN Crowd Dataset} \cite{111} It is also a publicly available dataset containing normal and abnormal crowd videos from University of Minnesota. Each video consists of an initial part of a normal behavior and ends with sequences of the abnormal behavior.

    \textbf{UCSD Anomaly Detection Dataset} \cite{112} This dataset was acquired with a stationary camera mounted at an elevation, overlooking pedestrian walkways. The crowd density in the walkways ranges from sparse to very crowded. Abnormal events are caused by either: (i) the circulation of non pedestrian entities in the walkways or (ii) anomalous pedestrian motion patterns.

    \textbf{Violent-Flows Dataset} \cite{113} It is a dataset of real-world video footages of crowd violence, along with standard benchmark protocols designed to test both violent/non-violent classification and violence outbreak detection. All the videos were downloaded from YouTube, and the average length of a video clip is 3.60 seconds.

    \textbf{CUHK Dataset} \cite{24} This dataset is for research on activity or behavior analysis in crowded scenes. It includes 2 subsets: a traffic dataset (MIT traffic) and a pedestrian dataset. The traffic dataset includes a traffic video sequence of 90 minutes long. Ground truths about pedestrians of some sampled frames are manually labeled. The pedestrian dataset were recorded in New York's grand center station, contain a 30 minutes long video sequence, without giving any ground truth or labeled data.

    \textbf{QMUL Dataset} \cite{115} This dataset has two subsets: the first contains three different dense traffic flow videos at crossroads, of nearly 60 minutes long; the second contains a video of shopping mall from a publicly accessible webcam. Over 60,000 pedestrians were labeled in 2000 frames, and the head positions of every pedestrians are labeled, making this dataset more convenient for crowd counting and profiling research.

    \textbf{PETS2009 Dataset} \cite{116} This dataset contains multi-sensor sequences of different crowd activities. It is composed of five parts: (i) calibration data, (ii) training data, (iii) person count and density estimation data, (iv) people tracking data, and (v) flow analysis and event recognition data. Each subset contains several sequences and each sequence contains different views (4 up to 8).

   \textbf{Rodriguez's Web-Collected Dataset} \cite{22} The dataset reported in was collected by crawling and downloading videos from search engines and stock footage websites (e.g., Gettyimages and YouTube).
    In addition to the large collection of crowd videos, the dataset contains ground-truth trajectories for 100 individuals, which were selected randomly from the set of all moving people. This dataset is not open to the public yet.

    \textbf{UCF Crowd Behavior Dataset} \cite{27} This dataset is collected from the web (Getty-Images, BBC Motion Gallery, Youtube, Thought Equity) and PETS2009, representing crowd and traffic scenes.
    It is publicly available in the form of image sequences. Differently from \cite{16}, it is mainly designed for crowd behaviors recognition, with ground-truth labels.

\section{Conclusions and Future Developments}\label{sec:conclusion}
    We have presented a review of the state-of-the-art techniques for crowded scene analysis across three key aspects: motion pattern segmentation, crowd behavior recognition and anomaly detection. The interested problems have become active research areas in recent decades because of their promising real-world applications.
    It can be seen that anomaly detection in crowded scenes has attracted quite a lot efforts, which reflects its importance in applications.
    Similar to these subtopics, feature representation is put as another important section deserving detailed descriptions, since feature representation, as an indispensable basis, is highly correlated with each of the three subtopics.
Furthermore, available knowledge of crowd from areas such as crowd dynamics is summarized beforehand. It could provide the fundamental crowd models for many scene analysis algorithms.

    Although a variety of representation approaches and models have been proposed, at the moment there is still no generally accepted solution to crowded scene analysis tasks.
    When retrospect the surveyed literatures, we can see two common models promising to visually characterize the crowded scenes: flow field model inspired by physics and generative topic model from machine learning.

    For flow field model, the crowd is treated as physical fluid and particles, and high density crowd behaves like a complex dynamic system. Many dynamical crowd evolution models have been proposed, along with the concepts of motion field and dynamical potential, borrowed from fluid dynamics community. By treating the moving crowd as a time dependent flow field which consists of regions with qualitatively different dynamics, the motion patterns emerging from the spatio-temporal interactions of the participants can be reflected.

    For topic model, the basic idea is ``a crowded scenes with its various events is a document with mixture of topics''. It could be along with different statistical assumptions. The topic model has the ability to automatically discover meaningful events or activities from the visual word co-occurrences. Moreover, it is flexible to utilize various features and other learning algorithms.

    Although a large amount of works has been done, many issues in crowded scene analysis are still open, and they deserve further research. In the following we list some of the promising topics.

    \textbf{Multi-Sensor Information Fusion} Crowded scenes often contain severe clutter and object occlusions, which are quite challenging for current visual-based techniques.
    To fuse information from multi-sensors is always an effective way to reduce the confusion and to improve the accuracy \cite{49}.
    Visual surveillance of crowded scenes could greatly benefit from the use of multiple sensors, such as audio, radar and laser.
    Multi-camera contexts could also be explored, as revealed by a recent work on group activity analysis~\cite{Zha2013}.
    By combining the multi-sensor data, different forms of information can complement to each other, and facilitate the system to obtain an accurate and comprehensive understanding of the scene.

    \textbf{Tracking-Learning-Detection Framework} Many current video analysis systems perform tracking, learning and detection by simple integration, without considering the interactions between the functional modules.
    To fully utilize the hierarchical contextual information, it is better for crowded scene analysis systems to simultaneously perform tracking, model learning, and behavior detection in a fully online and unified way.
    Some works on video surveillance have shown the advantages of the unified framework \cite{120,122}, which should require attention.

    The tracking module provides motion features, based on which crowd models are learned, and behaviors or events can be detected.
    On the other hand, crowded scene knowledge can facilitate accurate individuals tracking. For example, a person in particular crowd flow is always influenced by the global motions.
    Moreover, events or activities can be detected on the basis of tracking results and the learned models. They could also provide contextual knowledge for tracking and learning.
    A unified tracking-learning-detection framework can use all these contexts to improve these components simultaneously.

    \textbf{Deep Learning for Crowded Scene Analysis} Though various methods have been proposed on feature extraction and model learning in crowded scene analysis, still there is no public accepted crowded scene representation currently.
    In recent years, deep learning has achieved great success in several vision tasks related to visual surveillance and scene analysis ~\cite{DL:NIPS2012,Zeng2013Multi}. It has demonstrated its representation learning ability from multiple features. It also could be a prospective solution in crowded scene analysis, given enough training data. How to designate the framework and utilize the power of deep learning in tasks of crowded scene analysis deserve our future efforts.

    \textbf{Real-time Processing and Generalization} As such an area driven by practical applications, the real-time computation must be considered for the algorithms to work in real life.
    Current solutions usually target at accurate scene understanding, without considering the computation.
    Furthermore, many studies in the literature typically evaluate on video data of a specified condition.
    Research on the effective methods, which can well handle more generalized situations,
    could also be valuable.


\bibliographystyle{ieeetran}

\begin{IEEEbiography}[{\includegraphics[width=1in,height=1.25in,clip,keepaspectratio]{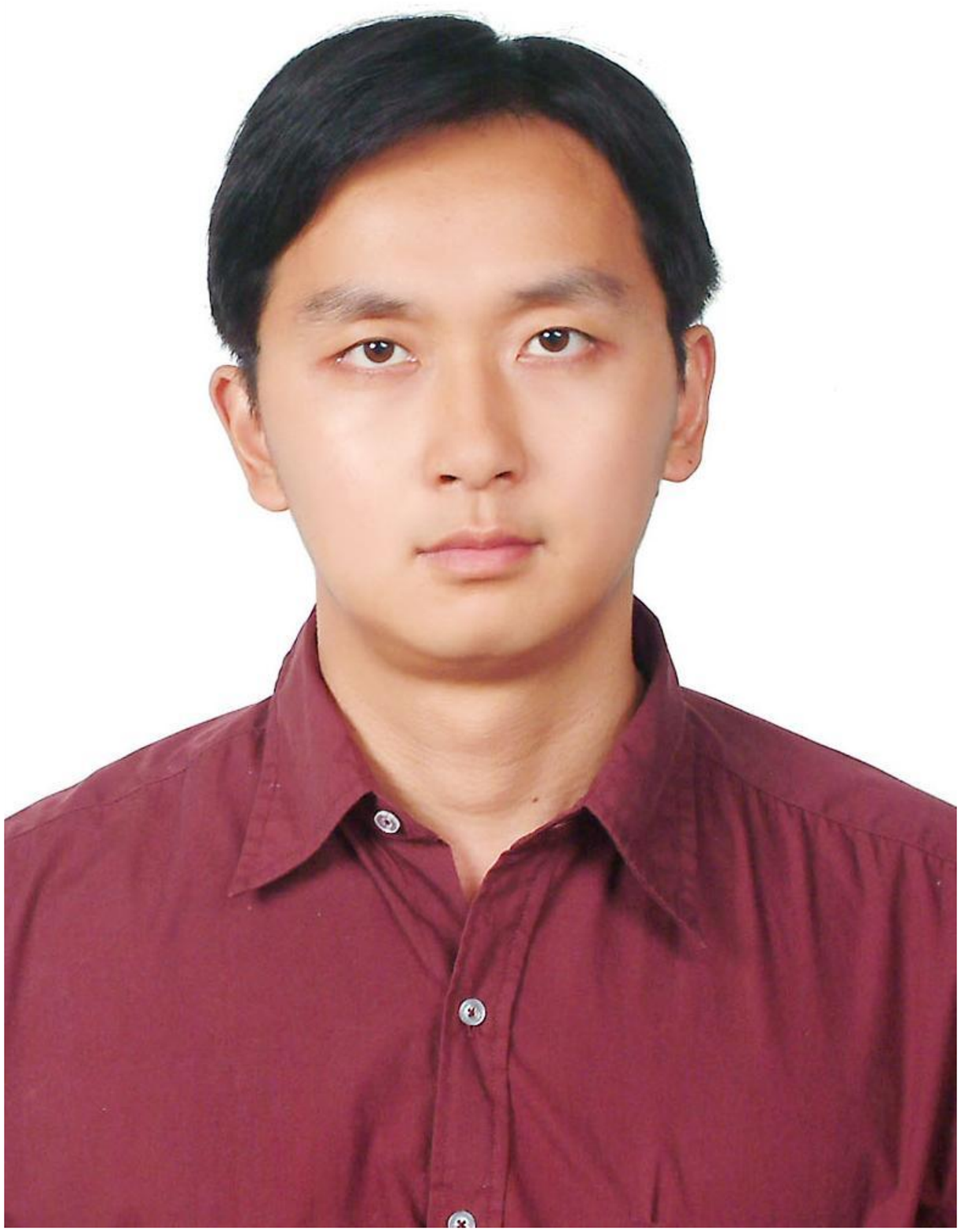}}]{Dr. Teng Li} received B.S. from University of Science and Technology of China (USTC) in 2001, M.S. from Institute of Automation, Chinese Academy of Sciences (CASIA) in 2004, and Ph.D. from Korea Advanced Institute of Science and Technolgy (KAIST) in 2010. He is currently a tenure-track professor with Anhui University. He received the Best Paper Award from IEEE T-CSVT in 2014, and the Best Paper Award of ICIMCS’09.
\end{IEEEbiography}
\begin{IEEEbiography}[{\includegraphics[width=1in,height=1.25in,clip,keepaspectratio]{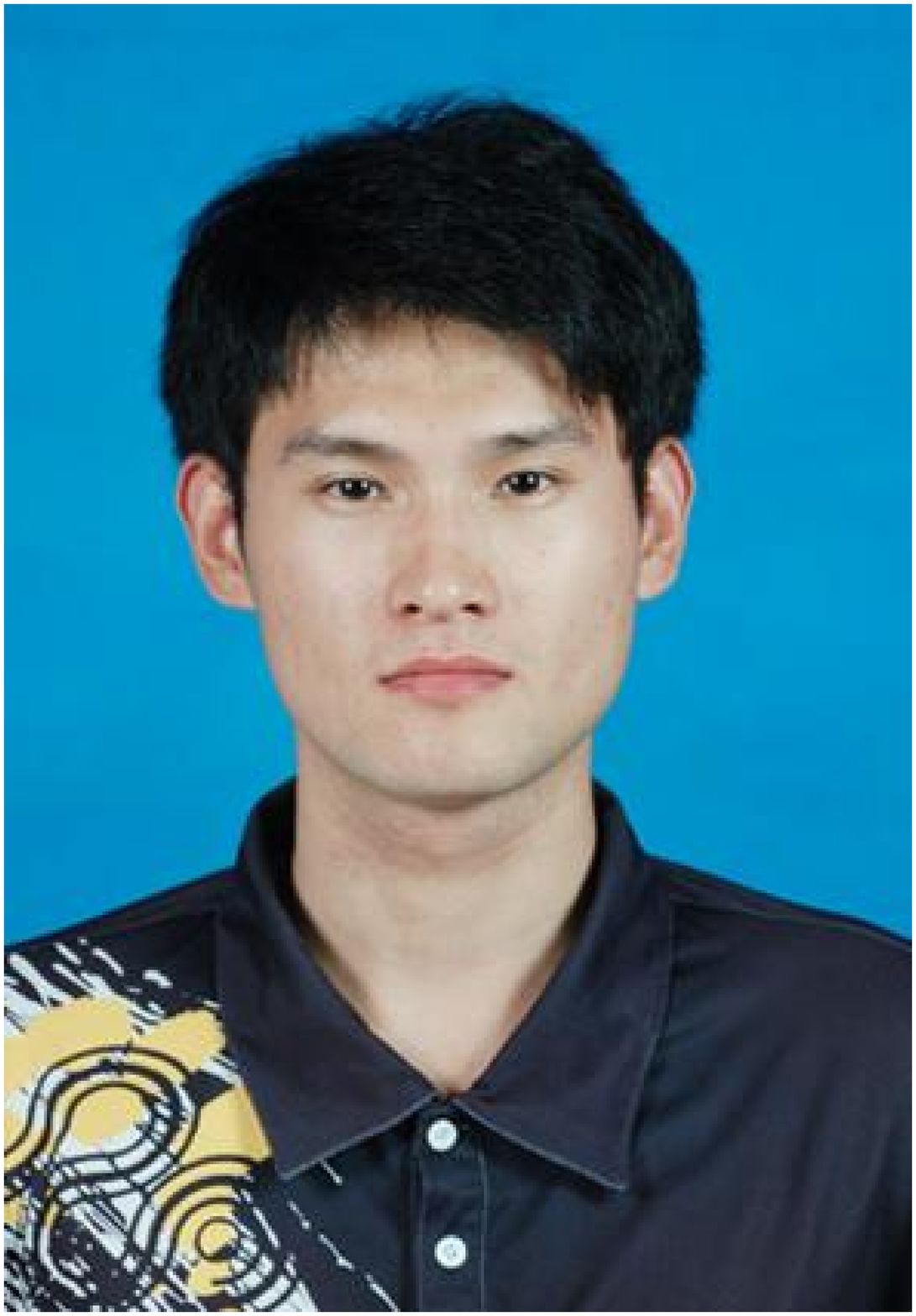}}]{Huan Chang} received B.S. from Anhui University (AHU), China in 2012. He is currently a master candidate with College of Electrical Engineering and Automation, Anhui University. His research interests are in the areas of computer vision.
\end{IEEEbiography}
\begin{IEEEbiography}[{\includegraphics[width=1in,height=1.25in,clip,keepaspectratio]{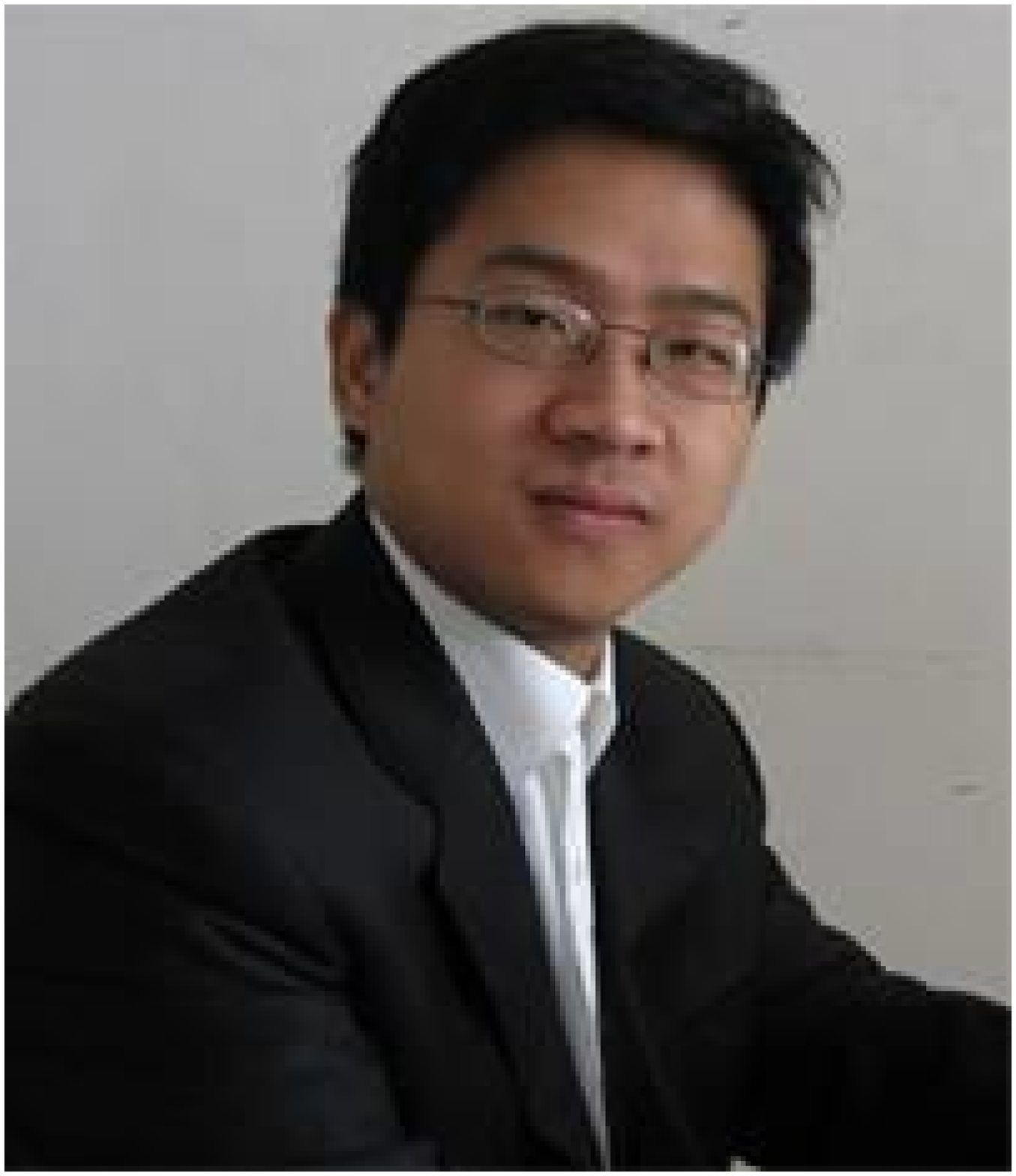}}]{Dr. Meng Wang} is a professor in the Hefei University of Technology, China. He received the B.E. degree and Ph.D. degree in the Special Class for the Gifted Young and the Department of Electronic Engineering and Information Science from the University of Science and Technology of China (USTC), Hefei, China, respectively.  His current research interests include multimedia content analysis, search, mining, recommendation, and large-scale computing. He received the best paper awards successively from the 17th and 18th ACM International Conference on Multimedia, the best paper award from the 16th International Multimedia Modeling Conference, the best paper award from the 4th International Conference on Internet Multimedia Computing and Service, and the best demo award from the 20th ACM International Conference on Multimedia.
\end{IEEEbiography}
\begin{IEEEbiography}[{\includegraphics[width=1in,height=1.25in,clip,keepaspectratio]{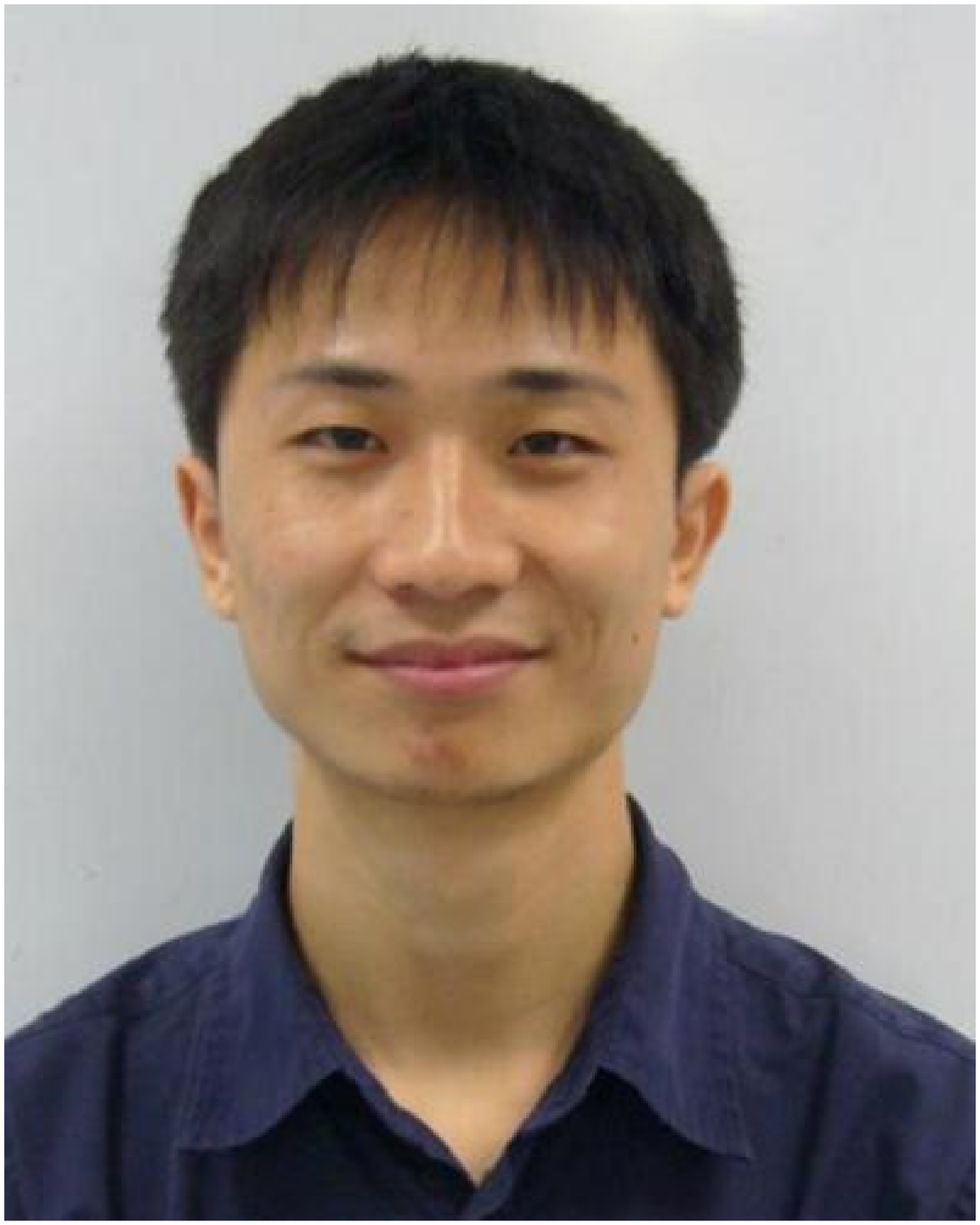}}]{Dr. Bingbing Ni} received his B.Eng. degree in Electrical Engineering from Shanghai Jiao Tong University (SJTU), China in 2005 and obtained his Ph.D. from National University of Singapore (NUS), Singapore in 2011. Dr. Ni is currently a research scientist in Advanced Digital Sciences Center, Singapore. His research interests are in the areas of computer vision, machine learning and multimedia. Dr. Ni worked in Microsoft Research Asia, Beijing as a research intern in 2009. He also worked as a software engineer intern in Google Inc., Mountain View, CA in 2010. He received the Best Paper Award from PCM’11 and the Best Student Paper Award from PREMIA’08. He won the first prize in International Contest on Human Activity Recognition and Localization (HARL) in conjunction with International Conference on Pattern Recognition, 2012. He also won the second prize in ChaLearn 2014 human action recognition challenge.
He is a member of the Association for Computing Machinery (ACM).
\end{IEEEbiography}
\begin{IEEEbiography}[{\includegraphics[width=1in,height=1.25in,clip,keepaspectratio]{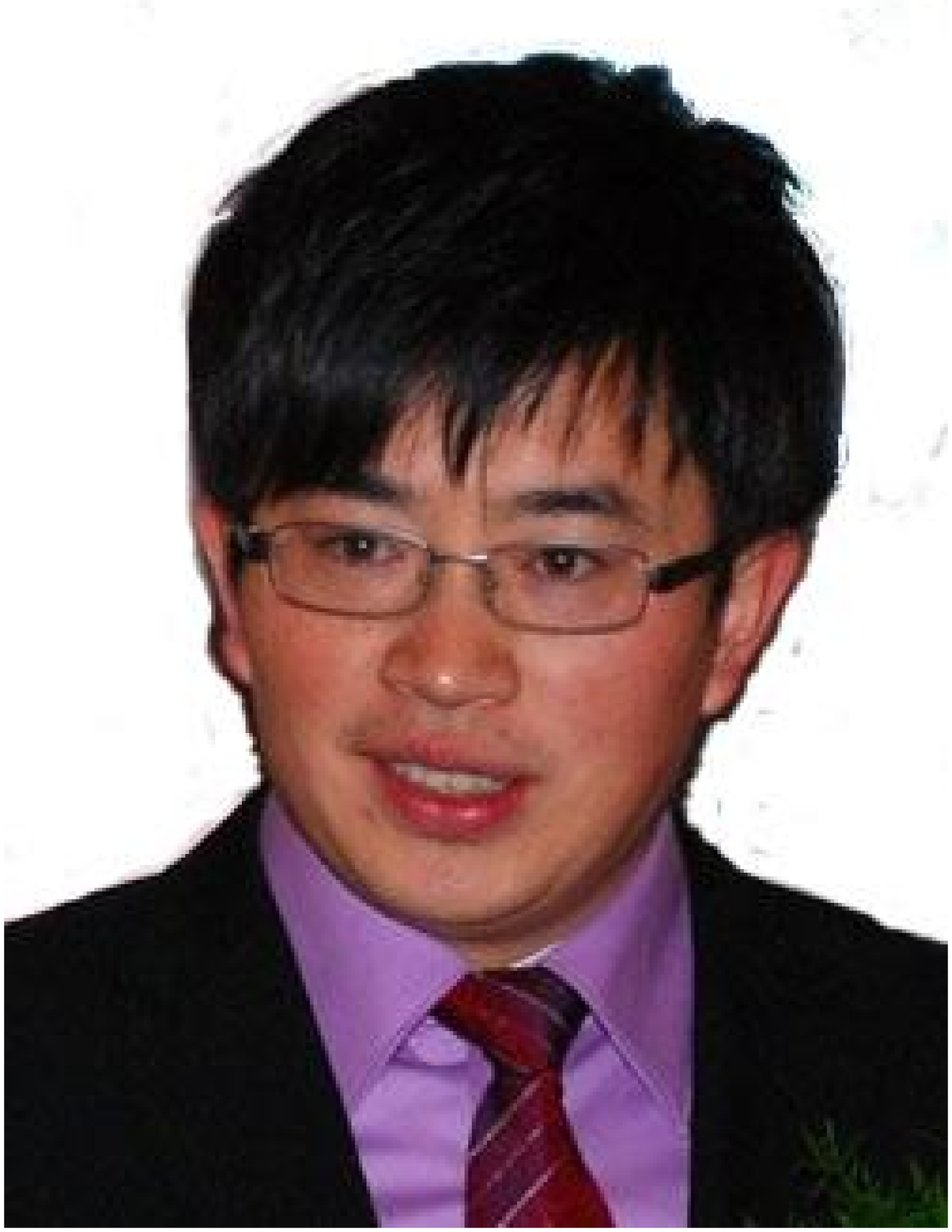}}]{Dr. Richang Hong} received the Ph.D. degree from the University of Science and Technology of China, Hefei, China, in 2008. He worked as a Research Fellow in the School of Computing, National University of Singapore, as a Research Fellow from September 2008 to December 2010.
He is now a Professor in Hefei University of Technology, Hefei, China. He has coauthored more than 60 publications in the areas of his research interests, which include multimedia question answering, video content analysis, and pattern recognition. Dr. Hong is a member of the Association for Computing Machinery. He was the recipient of the Best Paper Award in the ACM Multimedia 2010.
\end{IEEEbiography}

\begin{IEEEbiography}[{\includegraphics[width=1in,height=1.25in,clip,keepaspectratio]{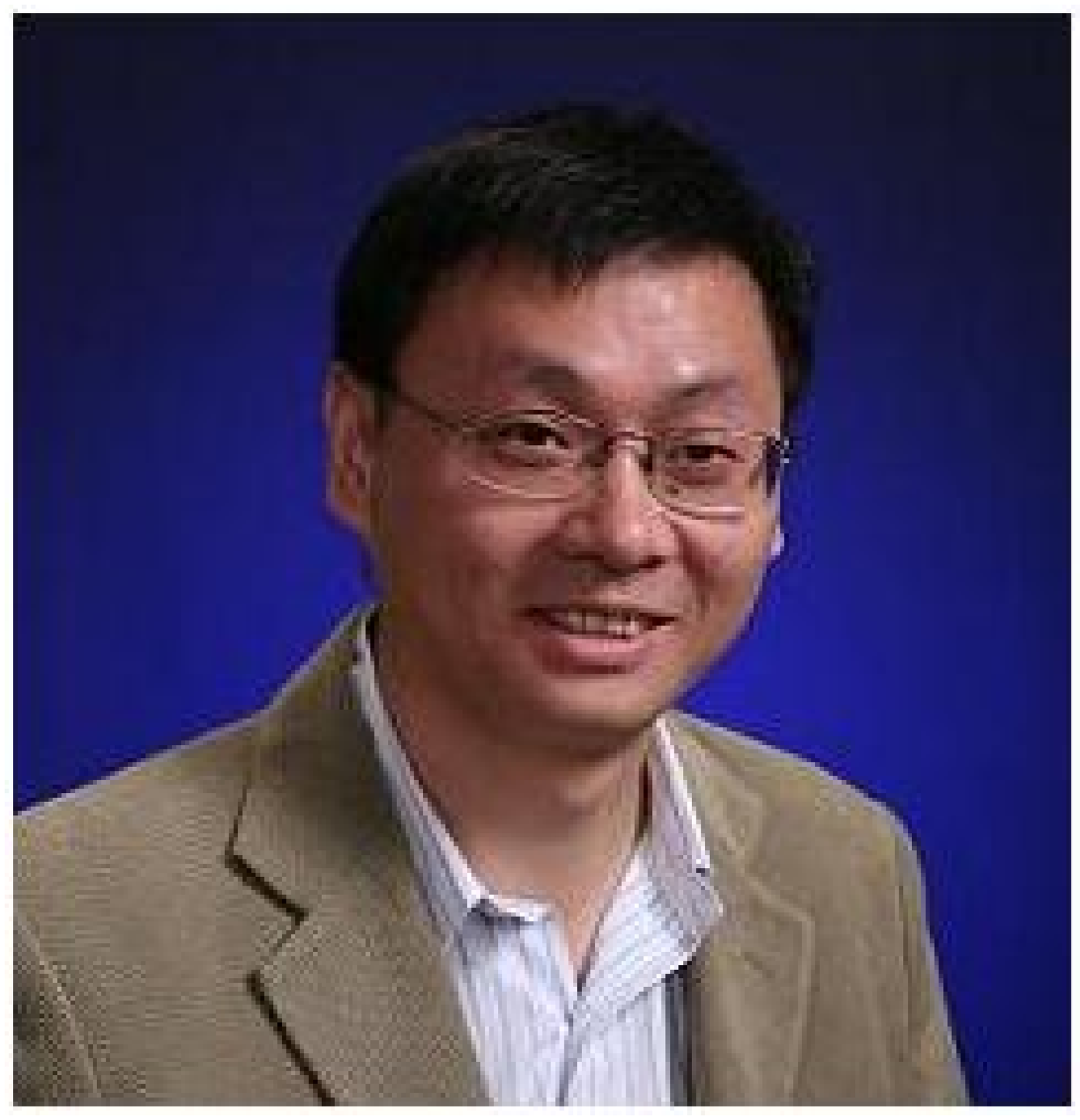}}]{Dr. Shuicheng Yan [SM'09]} is currently an Associate Professor in the Department of Electrical and Computer Engineering at National University of Singapore. Dr. Yan's research areas include computer vision, multimedia and machine learning, and he has authored/co-authored over 370 technical papers over a wide range of research topics, with Google Scholar citation more than 12,000 times and H-index-47. He received the Best Paper Awards from ACM MM13 (Best Paper and Best Student Paper), ACM MM’12 (demo), PCM'11, ACM MM’10, ICME’10 and ICIMCS'09, the winner prizes of the classification task in PASCAL VOC 2010-2012, the winner prize of the segmentation task in PASCAL VOC 2012, the honorable mention prize of the detection task in PASCAL VOC'10, 2010 TCSVT Best Associate Editor (BAE) Award, 2010 Young Faculty Research Award, 2011 Singapore Young Scientist Award, 2012 NUS Young Researcher Award, and the co-author of the best student paper awards of PREMIA'09, PREMIA'11 and PREMIA'12.
\end{IEEEbiography}

\end{document}